%% file: main.tex
\PassOptionsToPackage{dvipsnames}{xcolor}

\RequirePackage{amsmath,amssymb,amsfonts,amsthm,amstext}

\documentclass[table,a4paper]{ai2style/ai2}

\microtypesetup{expansion=false}

\definecolor{paperblue}{HTML}{285E9F}   
\definecolor{paperink}{HTML}{1D2733}    
\definecolor{paperbox}{HTML}{EBF2FA}    
\colorlet{ai2pink}{paperblue}
\colorlet{ai2accent}{paperblue}
\colorlet{ai2background}{paperbox}
\colorlet{ai2foreground}{paperink}
\colorlet{ai2foregroundDark}{paperink}

\usepackage{dsfont}
\usepackage{latexsym}
\usepackage{nicefrac}
\usepackage{array}
\usepackage{tabularx}
\usepackage{makecell}
\usepackage{threeparttable}
\usepackage{longtable}
\usepackage{pbox}
\usepackage{float}
\usepackage{wrapfig}
\usepackage{enumitem}
\usepackage{xspace}
\usepackage{xstring}
\usepackage{xargs}
\usepackage{pifont}
\usepackage{multicol}
\usepackage{xurl}
\usepackage{listings}
\usepackage{inconsolata}
\usepackage{algorithm}
\usepackage{algorithmic}
\usepackage{lineno}          
\usepackage{todonotes}       
\usepackage{tikz}
\usetikzlibrary{calc}
\usepackage[edges]{forest}
\tcbuselibrary{skins,theorems,breakable,fitting,hooks}

\usepackage{newunicodechar}
\newunicodechar{→}{\ensuremath{\rightarrow}}
\newunicodechar{Δ}{\ensuremath{\Delta}}

\setlist[itemize]{noitemsep}
\setlist[enumerate]{noitemsep}

\definecolor{hidden-draw}{RGB}{20,68,106}
\definecolor{hidden-pink}{RGB}{255,245,247}
\definecolor{boxcolor}{HTML}{d92523}
\definecolor{bulbcolor}{HTML}{e3b87f}
\definecolor{rliableolive}{HTML}{BBCC33}
\definecolor{rliableblue}{HTML}{77AADD}
\definecolor{rliablered}{HTML}{f63c44}
\definecolor{iquestblue}{HTML}{173C7F}
\definecolor{iquestazure}{HTML}{528FCC}
\definecolor{darkblue}{RGB}{84, 112, 198}
\definecolor{lightblue}{rgb}{0.85, 0.95, 1.0}
\definecolor{lightgreen}{rgb}{0.90, 1.0, 0.90}
\definecolor{lightorange}{rgb}{1.0, 0.95, 0.85}
\definecolor{lightpurple}{rgb}{0.95, 0.90, 1.0}
\definecolor{lightgray}{rgb}{0.97, 0.97, 0.97}
\definecolor{battery-empty}{rgb}{0.9, 0.9, 0.9}

\newcommand{\tablebold}[1]{\text{\bfseries #1}}

\newcommand{\taubench}{$\tau^2$-bench}

\newcommandx{\info}[2][1=]{\todo[linecolor=red,backgroundcolor=red!25,bordercolor=red,#1]{#2}}

\tcbset{
  aibox/.style={
    width=\linewidth, top=7pt, bottom=2pt,
    colback=rliablered!18!white, colframe=black, colbacktitle=black,
    enhanced, center,
    attach boxed title to top left={yshift=-0.1in,xshift=0.15in},
    boxed title style={boxrule=0pt,colframe=white,},
  }
}
\tcbset{
  aibox2/.style={
    width=\linewidth, top=7pt, bottom=2pt,
    colback=green!18!white, colframe=black, colbacktitle=black,
    enhanced, center,
    attach boxed title to top left={yshift=-0.1in,xshift=0.15in},
    boxed title style={boxrule=0pt,colframe=white,},
  }
}
\newtcolorbox{AIbox}[2][]{aibox,title=#2,#1}
\newtcolorbox{AIbox2}[2][]{aibox2,title=#2,#1}

\newcommand{\difficultybar}[1]{%
  \begin{tikzpicture}[baseline, scale=0.5, every node/.style={scale=0.8}]
    \foreach \i in {1,2,3,4,5} {
      \ifnum\i>#1
        \draw[fill=battery-empty] (\i*0.5-0.5, 0) rectangle (\i*0.5, 0.25);
      \else
        \pgfmathsetmacro{\colorlevel}{80 - 12*(\i)}
        \edef\x{\noexpand\draw[fill=blue!\colorlevel!white, opacity=0.9] (\i*0.5-0.5, 0) rectangle (\i*0.5, 0.25);}
        \x
        \draw[blue!50!black] (\i*0.5-0.5, 0) rectangle (\i*0.5, 0.25);
      \fi
    }
    \fill[battery-empty!70] (2.5, 0.08) rectangle (2.6, 0.17);
    \draw[battery-empty!70!black] (2.5, 0.08) rectangle (2.6, 0.17);
  \end{tikzpicture}%
}
\renewcommand{\arraystretch}{0.96}

\input{styles/visual_cases}

\hypersetup{
    pdftitle={Agentic On-Policy Distillation Implicitly Reshapes LLMs' ReAct Patterns},
    pdfpagemode=UseNone,
}

\let\cite\citep

\DeclareRobustCommand{\crest}[1]{\raisebox{-0.32em}{\includegraphics[height=1.3em]{#1}}}

\title{\color{black}\LARGE Agentic On-Policy Distillation Implicitly Reshapes LLMs' ReAct Patterns}

\authorOne{Deyang Kong\textsuperscript{1,*}\quad Zhaoyi Li\textsuperscript{2,*}\quad Sijian Ren\textsuperscript{3}\quad Ying Wei\textsuperscript{4}\quad Defu Lian\textsuperscript{2}\quad Shikun Zhang\textsuperscript{1}\quad Wei Ye\textsuperscript{1}}

\affiliation[\crest{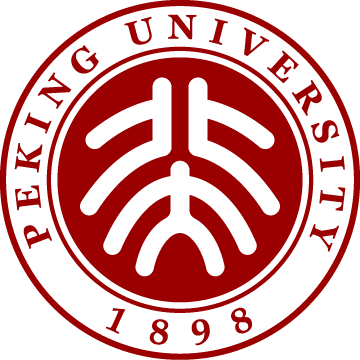}]{\textsuperscript{1}Peking University}
\affiliation[\crest{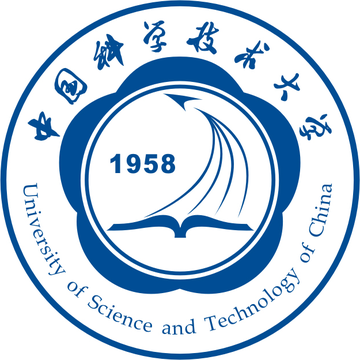}]{\textsuperscript{2}University of Science and Technology of China}
\affiliation[\crest{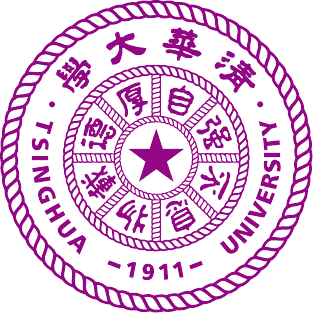}]{\textsuperscript{3}Tsinghua University}
\affiliation[\crest{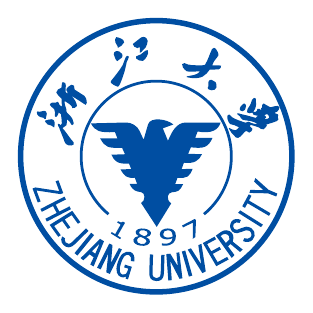}]{\textsuperscript{4}Zhejiang University}

\contribution[]{\textsuperscript{*}Equal Contribution}

\correspondence{\email{kong.deyang@foxmail.com}, \email{lizhaoyi777@mail.ustc.edu.cn}}

\abstract{On-policy distillation (OPD) transfers specialized capabilities to language-model agents, but its cross-domain effects on interaction remain unclear. We investigate how OPD reshapes ReAct interaction patterns across mathematics, ALFWorld, and $\tau^2$-bench, focusing on reasoning, actions, and observation acquisition. General-reasoning supervision reduces mathematical tool attempts from $70.4\%$ to $14.4\%$ while increasing accuracy from $16.7\%$ to $38.0\%$; code-interpreter supervision improves household actions and retail queries. In contrast, ALFWorld supervision reduces $\tau^2$-bench accuracy from $41.5\%$ to $6.6\%$, accompanied by frequent tool calls that violate user-information dependencies. To address negative transfer using only source-domain training trajectories, we explore selective OPD guided by teacher preference changes before and after domain-specific RL. Retaining the top $95\%$ of positions raises $\tau^2$ accuracy from $11.5\%$ to $29.9\%$ with a source-learning trade-off.
}

\begin{document}

\maketitle

\let\oldthefootnote\thefootnote

\input{sections_naacl/01_react_introduction}
\input{sections_naacl/02_react_related}
\input{sections_naacl/03_preliminaries}

\input{sections_naacl/04_react_patterns}
\input{sections_naacl/05_selecting_supervision}

\input{sections_naacl/06_react_discussion}
\input{sections_naacl/07_react_conclusion}
\label{page:main-end}

\section*{Limitations}
Our experiments focus on the Qwen3 model family with 4B students and mathematical and interactive tasks.
Extending the analysis to larger students, other model families,
and a broader range of agent environments would help establish
the generality of the observed transfer patterns.
The effects of supervision selection may also depend on the
training recipe and interaction protocol, motivating further
study across additional configurations.

\section*{Ethical Considerations}
This research helps identify potential risks arising when
specialized training changes an agent's interaction behavior.
In particular, changes in information gathering and user
confirmation can lead to actions based on unsupported assumptions
or actions taken before authorization is obtained.
Our findings can inform evaluations of tool-using agents and
encourage attention to feedback dependencies, confirmation
requirements, and execution permissions in practical applications.

\clearpage
\bibliography{references}

\clearpage
\beginappendix

\input{appendix_new/discussion}
\input{appendix_new/experimental_setup}
\input{appendix_new/evaluation_protocols}
\input{appendix_new/evaluation_results}
\input{appendix_new/analysis_preference_guided_selection}

\input{appendix_new/addition_controls}
\input{appendix_new/training_prompts}

\input{appendix_new/case_studies_complete}

\end{document}

%% file: styles/visual_cases.tex
\usepackage{tcolorbox}
\tcbuselibrary{breakable,skins}
\usepackage{listings}
\definecolor{opdgreen}{HTML}{287C6E}
\definecolor{opdblue}{HTML}{426B86}
\definecolor{opdrust}{HTML}{A95A48}
\definecolor{opdink}{HTML}{253B3B}
\newcommand{\opdrole}[1]{\par\medskip\noindent\textbf{\textcolor{opdgreen}{#1}}\par\nobreak\smallskip}
\newtcolorbox{opdcasebox}[2]{enhanced,breakable,
  colback=white,colframe=#2!65!white,colbacktitle=#2!9!white,
  coltitle=opdink,fonttitle=\small\bfseries,title={#1},
  boxrule=.45pt,arc=2pt,left=7pt,right=7pt,top=6pt,bottom=6pt,
  before skip=8pt,after skip=9pt,fontupper=\small\raggedright,
  pad at break*=2mm,title after break={#1 (continued)}}
\lstdefinestyle{opdcasecode}{basicstyle=\ttfamily\scriptsize,
  breaklines=true,breakatwhitespace=false,columns=fullflexible,
  keepspaces=true,showstringspaces=false,aboveskip=4pt,belowskip=4pt,
  literate={’}{{'}}1 {‘}{{'}}1 {“}{{``}}2 {”}{{''}}2
           {—}{{---}}3 {–}{{--}}2 {…}{{...}}3 {→}{{$\to$}}1}
\newcommand{\opddeltatok}[3]{\begingroup\setlength{\fboxsep}{.3pt}%
  \colorbox[HTML]{#1}{\strut\ifnum#2=1\relax\underline{#3}\else#3\fi}\endgroup\allowbreak}

%% file: sections_naacl/01_react_introduction.tex
\section{Introduction}

Large Language Model (LLM) agents' competence depends on how it connects reasoning, action, and observation. Mathematical assistants may call python interpreter~\citep{xue2026simpletir} to calculate or directly derive an answer; Service agents may query an account or request a missing identifier. Reasoning and Acting (ReAct)~\citep{yao2022react}, which interleaves reasoning traces with task-specific actions, allowing an LLM to plan, track its progress, and handle exceptions through thought while gathering information from external environments such as search APIs or tools, decouples reasoning with actions and observations and becomes the most widely-used paradigm~\citep{schick2023toolformer,qin2024toolllm} which LLMs use to solve agentic tasks~\citep{jimenez2024swebenchlanguagemodelsresolve,wei2025browsecompsimplechallengingbenchmark}. On-Policy Distillation (OPD), which transfers teacher capabilities by supervising trajectories sampled by the student~\citep{agarwal2024policy,song2026survey}, has also become an important step in LLM post-training pipeline~\citep{ma2026mopd,team2026kimi,xu2026deepseek}. OPD training can affect models' choices, especially in the multi-turn agentic tasks, including how to organize reasoning and tool-use, how to obtain new information, interact with the environment, and so on.


We study these ReAct interaction patterns across four domains: (1) pure natural-language general reasoning~\citep{hendrycks2021measuring} (GR), (2) general reasoning with code-interpreter tools~\citep{heyueya2023solvingmathwordproblems,qian2025toolrlrewardtoollearning} (CI), (3) ALFWorld~\citep{ALFWorld20}, and (4) $\tau^2$-bench~\citep{barres2025tau}. We find two observable ReAct pattern shifts in trajectories of the above agentic tasks. First, OPD can shift the route to a mathematical answer between tool-assisted computation and direct natural-language reasoning (\S~\ref{sec:dimensions}): GR-domain OPD training starkly reduces models' tool attempt rate from $70.4\%$ to $14.4\%$ in the CI tasks. Second, it can change how the student chooses actions, adjusts them after environment feedback, and waits for information or authorization before acting (\S~\ref{sec:dependencies}): CI supervision improves both ALFWorld and $\tau^2$ tasks, accompanied by fewer repeated actions and more complete queries. Meanwhile, we find that ALFWorld-domain OPD makes student's $\tau^2$ performance drastically drop from 41.5\% to 6.6\%. The student frequently generates request-like text alongside tool calls without yielding the turn to the user. More importantly, semantic review identifies calls that depend on unavailable user information or authorization in $55.8\%$ of trajectories. These findings connect cross-domain outcomes to reasoning routes, feedback-driven actions, and the acquisition of observations.

\begin{wrapfigure}{r}{0.45\textwidth}
\centering
{\includegraphics[width=\linewidth]{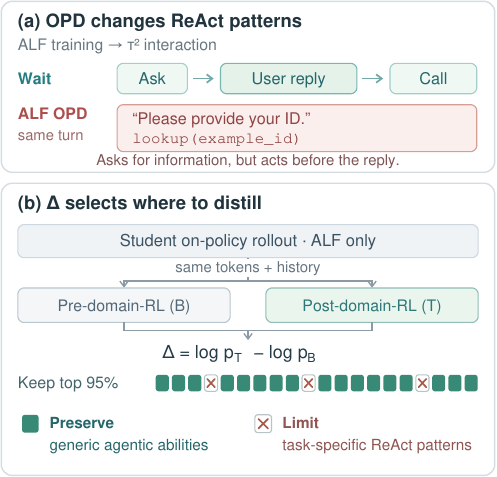}} \caption{Agentic OPD can change the ReAct patterns of LLM agents. Our proposed mitigation approach helps alleviate the overfitting to task-specific patterns.}
\label{fig:teaser}
\end{wrapfigure}

We analyze that severely overfitting to \textbf{\textit{task-specific ReAct patterns}} is the main cause of the observed negative transfer. Hence, we design a selective policy (\S~\ref{sec:preference-selection}), which keeps the generic agentic abilities (mostly derived from the agentic RL training~\citep{yu2025demystifying}) and discards the plain transfer of surface-level task-specific ReAct patterns, as illustrated in Figure~\ref{fig:teaser}. Given a student's rollout, we use both of the teacher's checkpoints before and after domain-specific RL to compute token log-probabilities for sampled tokens. Naturally, the difference ($\Delta$) of log-likelihood between two checkpoints (post-/pre-RL) indicates how agentic RL changes teacher models' preference on the target token. We find that excluding a small fraction of tokens (i.e., with the least $5\%$ RL-induced $\Delta$).) in the ALFWorld-domain OPD training can effectively mitigate the catastrophic negative transfer on the $\tau^2$ tasks, while maintaining most of the performance gain on ALFWorld tasks. We also extend the analysis to a RL-trained GR teacher: selective supervision preserves more CI tool use at the same mean GR accuracy, with a reduction in CI accuracy.


Our key contributions are two-fold: (1) We systematically examine how OPD training, particularly from domain-specialized teachers, changes the reasoning, action, and observation-acquisition patterns of LLM agents across math reasoning (with and without Python code interpreter), ALFWorld, and $\tau^2$-bench. We identifies both positive cross-domain transfer and negative transfer, notably ALF-domain OPD training will drastically make student's $\tau^2$ performance drop down; and (2) To mitigate this failure using only ALF-source training trajectories, we propose selecting OPD token positions according to the log-probability change between a teacher checkpoint before and after domain-specific RL, excluding the lowest-tail, which largely alleviates the degradation of $\tau^2$ accuracy while maintaining most of the performance gain on the ALFWorld. GR-to-CI extension examines whether the method also preserves mathematical tool-use behavior.

%% file: sections_naacl/02_react_related.tex
\section{Related Work}
\paragraph{On-Policy Distillation.}


Knowledge distillation transfers teacher behavior to students \citep{hinton2015distilling}. OPD supervises student-generated contexts \citep{agarwal2024policy,song2026survey}, while multi-teacher OPD integrates domain experts \citep{ma2026mopd}. Recent work explores entropy-aware KL objectives \citep{jin2026entropy}, OPD optimization geometry \citep{shen2026geometry}, and efficient offline distillation \citep{wu2026lightning}. For agentic OPD, SEED distills hindsight skills \citep{wu2026seed}, ATOD introduces turn-aware weighting \citep{tan2026atod}, and ActFirst-OPD separates action and reasoning \citep{zheng2026act}. Studies further examine teacher--student mismatch in multi-turn agents \citep{wang2026tcod,zhong2026sod,zhou2026sage} and cross-domain transfer \citep{li2026every,wang2026beyond}. Delta-based distillation leverages teacher--predecessor preference differences \citep{heo2026policy}. We examine these effects in agentic trajectories~\citep{shen2026top}, tracing solution routes, responses to feedback, and the timing of information requests. This behavioral focus complements work on forgetting \citep{kirkpatrick2017overcoming,lopez2017gradient} and multi-task interference \citep{yu2020gradient}.

\paragraph{LLM Agents' ReAct Patterns.}
ReAct couples reasoning with actions and observations \citep{yao2022react}; Toolformer and ToolLLM develop tool-use capabilities \citep{qin2024toolllm,schick2023toolformer}, and API-Bank evaluates tool-augmented interaction \citep{li2023api}. Reasoning methods influence the route to an answer through intermediate derivations and sampled solutions \citep{wei2022chain,kojima2022large,wang2022self,xu2025toward}. Dynamic reasoning strategies \citep{aoki2024first} and feedback-based revision \citep{shinn2023reflexion} likewise make process behavior important. Our comparisons ask how such observable patterns change after domain distillation, including when improved answer accuracy accompanies less tool use and when frequent actions bypass necessary user feedback. Tool uncertainty \citep{liu2024uncertainty}, procedurally generated environments \citep{sullivan2025procedural}, and prompt-injection evaluation \citep{debenedetti2024agentdojo} address complementary aspects of reliable interaction.

%% file: sections_naacl/03_preliminaries.tex
\section{Preliminaries and Experiment Settings}
\label{sec:setup}

\paragraph{Agent interaction.}
An agent receives a goal $x$ and interacts with external participants and systems. At decision turn $k$, its history $h_k$ contains the instructions, previous assistant outputs, tool results, environment observations, and user replies. The policy generates an output $u_k\sim\pi_\theta(\cdot\mid h_k)$. An output may communicate with a user, invoke a tool, or complete the task. The interaction process then supplies feedback $o_{k+1}$, producing $h_{k+1}=h_k\mathbin{\Vert}u_k\mathbin{\Vert}o_{k+1}$. For an assistant-generated token $y_t$ at decision turn $k$, we denote its complete autoregressive context by $H_t$.

We examine \emph{ReAct interaction patterns} by analyzing how agents organize reasoning, acting, and obtaining observation throughout an interaction~\citep{yao2022react}. Reasoning can derive an answer. Acting selects an operation and its arguments. Obtaining observations includes requesting information and yielding the turn so another participant can provide it.
They provide three observable perspectives on a trajectory.

\paragraph{OPD and MOPD.}
Let $x$ denote a prompt sampled from the training dataset $\mathcal{D}$, $\pi_{\theta}$ and $\pi_{\phi}$ the student and teacher policies, $y=\{y_t\}^T_{t=1}$ a student-generated response, and $h_t=(x,y_{<t})$ the context at step $t$. On-Policy Distillation (\textbf{OPD}) samples \textbf{on-policy} trajectories from the student policy $\pi_{\theta}$ and obtains dense token-level supervision from the teacher policy $\pi_{\phi}$. The student is optimized by minimizing $\mathcal{L}_{\mathrm{OPD}}(\theta)$, the reverse KL divergence between the student and teacher~\citep{gu2024minillm}:

{\small
\begin{align*}
\mathop{\mathbb{E}}\limits_{\substack{
    x\sim\mathcal{D} \\
    y\sim\pi_{\theta}(\cdot\mid x)
}}
\left[
    \frac{1}{T}
    \sum_{t=1}^{T}
    D_{\mathrm{KL}}
    \left(
        \pi_{\theta}(\cdot\mid h_t)
        \,\Vert\,
        \pi_{\phi}(\cdot\mid h_t)
    \right)
\right].
\end{align*}
}
In practice~\cite{xiao2026mimo,zeng2026glm, team2026kimi}, the reverse KL is estimated by a sampled-token $k_1$ approximation~\citep{schulman2020approximating}, so that $\mathcal{L}_{\mathrm{OPD}}(\theta)$ can be written as a reinforcement-learning objective (i.e., \textbf{Policy-Gradient (PG) style OPD}):

{\small
\begin{align*}
\mathcal{L}^{\mathrm{PG}}_{\mathrm{OPD}}(\theta)
&=
-
\mathbb{E}_{x,y}
\left[
    \frac{1}{T}
    \sum_{t=1}^{T}
    \widehat{A}^{\mathrm{OPD}}_t
    \log\pi_{\theta}(y_t\mid h_t)
\right],\\
\widehat{A}^{\mathrm{OPD}}_t
&=
\operatorname{sg}
\left[
    \log\pi_{\phi}(y_t\mid h_t)
    -
    \log\pi_{\theta}(y_t\mid h_t)
\right],
\end{align*}
}
where $\operatorname{sg}[\cdot]$ denotes the stop-gradient. We adopt PG-style OPD throughout: a token is reinforced when the teacher assigns it higher probability than the student, and suppressed otherwise. Multi-Teacher On-Policy Distillation (\textbf{MOPD})~\citep{ma2026mopd} is a natural extension of OPD and now a standard post-training paradigm~\citep{xiao2026mimo,team2026kimi} for integrating capabilities from multiple domains: the student samples from its own rollouts, \textbf{each prompt is routed to its corresponding domain teacher}, and the optimization procedure is identical to single-teacher OPD.

\paragraph{Experimental setup.}
In this work, we study four representative task domains with distinct ReAct patterns: (1) single-turn \textbf{General Reasoning (GR)} task, which uses natural language to solve math problems without calling tools; (2) \textbf{Code-Interpretation (CI)}~\citep{xue2026simpletir}, using \texttt{Python} code-interpreter to solve complex math problems; (3) \textbf{ALFWorld (ALF)}~\citep{ALFWorld20}, which simulates real-world embodied agentic tasks and supplies text-based household action feedback; and (4) \textbf{\taubench{} ($\mathbf{\tau^2}$)}~\citep{barres2025tau} adds simulated users and account-side tools in airline, retail, and telecom tasks. Student models for all of the OPD experiments are Qwen3-4B-Instruct-2507~\citep{yang2025qwen3}.

For evaluation, we use AIME2026 problems~\cite{aime26} sampled 16 times and BeyondAIME problems~\cite{bytedance_seed_2025_beyondaime} sampled five times in each mathematical setting. ALFWorld includes 140 in-distribution and 134 out-of-distribution tasks sampled four times. $\tau^2$-bench includes 100 tasks across airline, retail, and telecom, each sampled eight times. The OPD teacher models are derived through applying domain-specific RL training on Qwen3-4B-Instruct-2507 and Qwen3-30B-A3B-Instruct-2507. The experimental settings are concluded in Table~\ref{tab:settings}. Training prompts and tool interfaces are provided in Appendix~\ref{app:training-prompts}.

We compare single-source students, four pairwise compositions, and a CI/ALF/$\tau^2$ composition.
Model labels indicate training sources, with each model evaluated at a fixed checkpoint selected during source-domain validation. Large rollout batches support substantial learning within 50--60 updates for comparisons; the learning curves and checkpoint identities are given in Appendix~\ref{app:repro-checkpoints}. Evaluation protocols are specified in Appendix~\ref{app:evaluation}.

%% file: sections_naacl/04_react_patterns.tex
\section{Agentic OPD Reshapes ReAct Patterns}
\label{sec:patterns}

\input{figures/overview}

We train students with OPD on individual task domains and with MOPD on mixtures of two or three domains, then evaluate each student across tool-enabled and no-tool mathematics, ALFWorld, and $\tau^2$-bench. Figure~\ref{fig:overview} shows that training on one task can substantially change performance on other tasks. For example, CI-only OPD \textbf{\textit{observably raises}} ALFWorld accuracy from 0.129 to 0.402 ($+0.273$) and $\tau^2$ accuracy from 0.415 to 0.510 ($+0.095$). ALF-only OPD instead \textbf{\textit{drastically lowers}} $\tau^2$ accuracy from 0.415 to 0.066 ($-0.349$). The effects of domain-specific supervision therefore extend beyond the task that supplies its training trajectories, with different outcomes across source--target pairs.

Task scores establish the scope of this cross-domain effect but do not show what changed in the student's interaction policy. We examine two observable ReAct patterns in the resulting trajectories. First, OPD can shift the route to a mathematical answer between tool-assisted computation and direct natural-language reasoning (\S~\ref{sec:dimensions}). Second, it can change how the student chooses actions, adjusts them after environment feedback, and waits for information or authorization before acting (\S~\ref{sec:dependencies}). These analyses connect the performance differences in Figure~\ref{fig:overview} to specific decisions in the agent's reasoning and interaction process.

\subsection{Reasoning Patterns Shift}
\label{sec:dimensions}

\begin{wraptable}{r}{0.45\textwidth}
\caption{The statistics on CI math tasks.}
\label{tab:reasoning}
\centering\footnotesize\setlength{\tabcolsep}{3pt}
\input{tables/reasoning}
\end{wraptable}

Table~\ref{tab:reasoning} partitions correct mathematical outputs by whether they include a tool attempt. Following GR training, correct outputs without tools account for 30.1\% of all trajectories, compared with 2.4\% for Base. Correct outputs containing a tool attempt decrease from 14.3\% to 7.9\%. The gain in total accuracy therefore accompanies a substantial change in the route by which correct answers are produced.

Code features help locate this change. Trajectories containing loop-like code decrease from 304 to 97, and those containing solver calls decrease from 75 to eight. In a matched-input urn problem, Base increments the total number of marbles and enumerates the red count until two hypergeometric probabilities agree. The GR student instead cancels the binomial coefficients, derives $b=3k+2$, $r=5k+4$, and $n=8k+6$, and applies the lower bounds on both colors. Both obtain the correct answer, 190. Further cases appear in Appendix~\ref{app:cases}.

ALF training exhibits a related shift in mathematics. Its accuracy is 38.3\%, tool-attempt rate 11.5\%, and correct no-tool share 30.6\%. In a rotation-geometry case, Base constructs the wrong triangle and uses Python to calculate the resulting expression. The ALF student derives the rotated point in coordinates and obtains the correct answer without a tool. By contrast, the CI student retains a 73.0\% tool-attempt rate, and correct tool-containing outputs account for 47.6\% of its 52.4\% accuracy.

\begin{tcolorbox}[title={Observation 1: OPD shifts reasoning routes},
colback=blue!2,colframe=blue!65,fonttitle=\small\bfseries,boxrule=.4pt]
\small In tool-enabled mathematics, GR and ALF supervision shift the student models' reasoning route toward \textbf{\textit{natural-language derivations}} (e.g., leveraging sophisticated algorithms to avoid large computational loads), while CI supervision favors \textbf{\textit{solutions containing \texttt{Python} code tool-calls}} (e.g., using more loop code for brute-force search). OPD therefore changes the student's route to an answer across domains.
\end{tcolorbox}

\begin{table*}[t]
\caption{Positive transfer from CI. (a) Calls are complete generated blocks per rollout; Repeat is exact repetition among adjacent actions; Change is a changed action after no-op feedback with a following action. (b) All/Retail are overall and retail task accuracy; Tool is rollout-level attempt coverage; Error is the fraction of rollouts with error feedback; Query is complete reference-query coverage on the 200 eligible retail rollouts. Calls again use a per-rollout denominator. Detailed metric definitions and denominators are provided in Appendix~\ref{app:evaluation}.}
\label{tab:positive}
\centering\small
\begin{minipage}{.43\textwidth}
\centering\textbf{(a) CI $\rightarrow$ ALFWorld}\par\smallskip
\setlength{\tabcolsep}{4pt}\input{tables/positive_alf}

\end{minipage}\hfill
\begin{minipage}{.55\textwidth}
\centering\textbf{(b) CI $\rightarrow$ $\tau^2$}\par\smallskip
\setlength{\tabcolsep}{3.5pt}\input{tables/positive_tau}
\end{minipage}
\end{table*}

\subsection{Action and Observation Patterns Shift}
\label{sec:dependencies}
\paragraph{Action: CI-OPD enables more productive environment interactions in ALFWorld and $\tau^2$ tasks.}
In ALFWorld, Table~\ref{tab:positive}a shows that accuracy rises to 0.402 while adjacent action repetitions fall from 4.5\% to 0.7\%. After a ``Nothing happens'' observation, the fraction of subsequent actions that change increases from 90.8\% to 94.3\%. The mean number of calls rises from 12.97 to 18.94, so the improvement combines longer interaction with fewer consecutive repetitions.
For an instance, base student model stops after searching empty drawers in a bowl-on-coffee-table task. The CI-trained student continues to the dining tables and finds the bowl. It first tries placing it without moving, then responds to no-op feedback by moving to the coffee table and completing the task.

CI training also raises $\tau^2$ task accuracy  to 0.510. As shown in Table~\ref{tab:positive}b, the overall fraction of rollouts containing a tool attempt is nearly unchanged, 95.9\% versus 95.8\%. Retail contributes 55 of the 76 additional successful samples, with accuracy increasing from 0.481 to 0.653. On retail tasks with reference lookup actions, the share of trajectories that generate all reference queries increases from 54.5\% to 67.5\%. The fraction of all $\tau^2$ rollouts with error feedback falls from 32.6\% to 23.5\%. These gains accompany more complete information retrieval and fewer error-containing trajectories. For a payment-method change, Base assumes the order already uses the requested card without retrieving it. The CI-trained student queries the order, observes a different card, checks the gift-card balance, obtains confirmation, and successfully changes payment. CI also produces more mixed communication--tool turns: improved query sequences and turn protocol compliance are different properties.

\begin{wraptable}{r}{0.38\textwidth}
\caption{Information acquisition in $\tau^2$ interactions. Accuracy measures task success. }
\label{tab:timing}
\centering\footnotesize\setlength{\tabcolsep}{3pt}
\input{tables/semantic_timing}

\end{wraptable}
\paragraph{Action: OPD changes how agents obtain information.}
In contrast to CI's positive transfer, ALF-only OPD reduces $\tau^2$ task accuracy from 41.5\% to 6.6\%. We discover that this decline accompanies a change in how the student obtains information from users.

We measure ask--call co-occurrence as the fraction of assistant outputs containing detected request-like text that also contain a generated tool call. As shown in Table~\ref{tab:timing}, this rate reaches 94.3\% for the ALF student, indicating that asking and calling frequently occur within the same turn. Such text is not delivered to the user, and co-occurrence may indicate a problem when the call requires information not yet provided by the user.

To distinguish these cases, we additionally report the \emph{dependency-violation rate}: the fraction of trajectories containing at least one call that depends on user information or authorization requested in the same output but not yet obtained. These violations are identified through LLM-based semantic review of the interaction history available before each call (Appendix~\ref{app:semantic}). As shown in Table~\ref{tab:timing}, dependency violations occur in 55.8\% of ALF student trajectories, compared with 0.50\% for Base. These results indicate that the ALF-trained student frequently proceeds with tool calls before obtaining necessary user information or authorization. Figure~\ref{fig:dialogue} illustrates this pattern. The ALF student generates a request for a user ID but immediately calls the lookup tool using an example ID from the tool schema, without waiting for a reply. The ALF+$\tau^2$ student instead waits for the user's ID before making the corresponding call. This contrast is consistent with its much lower ask--call co-occurrence and absence of detected dependency violations.

\begin{figure}[bt]
\input{figures/dialogue}
\caption{Excerpts from trajectories for the same annotated task. The ALF student generates a user-ID request alongside a tool call using a schema-example ID. ALF+$\tau^2$ waits for the user's actual ID.}
\label{fig:dialogue}
\end{figure}

Notably, the ALF teacher rarely exhibits ask--call co-occurrence on $\tau^2$, so the student's pattern does not simply mirror the teacher's target-domain behavior. ALFWorld typically supplies observations through environment actions, whereas $\tau^2$ also requires waiting for user-provided information or approval. The observed transfer is therefore consistent with a shift toward tool-mediated observation acquisition, even when the required feedback must come from the user.

\begin{tcolorbox}[title={Observation 2: OPD changes Action Patterns},
colback=blue!2,colframe=blue!65,fonttitle=\small\bfseries,boxrule=.4pt]
\small Domain-specific OPD supervision (e.g., from CI) can be associated with \textbf{\textit{more effective and productive action patterns}} after environment feedback, which can also generalize to other held-out agentic tasks. However, domain-specific OPD supervision (e.g., from ALF) can instead \textbf{\textit{starkly change the action habit of obtaining information}} by LLM agents (e.g., producing frequent tool-calls rather than requesting feedback from the user-side in the $\tau^2$ tasks).
\end{tcolorbox}

\subsection{Task-matched supervision in MOPD}
\label{sec:mopd-observation}
The pronounced shifts above arise after single-source training; MOPD tests whether adding supervision from the evaluated task domain changes these observations. Table~\ref{tab:composition} compares four pairwise compositions and the three-teacher student through scores and conversational behavior.

\begin{table}[ht]
\caption{Teacher composition. (a) Accuracy on CI/GR, ALFWorld, and $\tau^2$. (b) Ask--call uses request turns; error repetition uses post-error turns. Definitions and counts appear in Appendix~\ref{app:evaluation}.}
\label{tab:composition}
\centering\small\setlength{\tabcolsep}{3.5pt}
\begin{minipage}[t]{0.57\textwidth}
\centering
\textbf{(a) Four-domain accuracy}\par\smallskip \footnotesize\setlength{\tabcolsep}{2.5pt}
\input{tables/composition_scores}
\end{minipage}\hfill
\begin{minipage}[t]{0.40\textwidth}
\centering
\textbf{(b) Conversational timing}\par\smallskip \footnotesize\setlength{\tabcolsep}{2pt}
\input{tables/composition_behavior}
\end{minipage}
\end{table}

(1) In mathematics, CI+GR reaches 51.3\% accuracy with 74.4\% tool attempts when tools are available, while achieving 44.5\% accuracy in the no-tool setting. This joint behavior contrasts with GR-only training, which reaches 38.0\% accuracy and 14.4\% tool attempts in tool-enabled mathematics. The route shift observed after GR-only supervision is thus not fixed when CI supervision is also present. (2) Adding $\tau^2$ prompts and their specialist teacher to ALF training raises $\tau^2$ accuracy from the ALF-only student's 0.066 to 0.689. Same-turn ask--call co-occurrence falls from 94.3\% to 0.24\%, and the reviewed ALF+$\tau^2$ rollouts contain no current-request dependency violations. At the same time, ALFWorld accuracy remains high (0.834 versus 0.855 for ALF-only). Within these evaluated tasks, task-matched supervision largely removes the specific harmful timing pattern identified in \S\ref{sec:dependencies}.

However, \textbf{\textit{coverage by task-matched teachers is nevertheless limited to the training sources.}} ALF+CI has no $\tau^2$ supervision: its $\tau^2$ performance still \textbf{drops by $\mathbf{7\%}$}, and meanwhile \textbf{ask--call co-occurrence remains $\mathbf{25.6\%}$}. This held-out result shows that combining teachers can still leave ReAct pattern (negative) transfer in an unseen task domain. Training cannot enumerate every agentic task or feedback protocol. The observed recovery therefore motivates more principled methods for preserving feedback dependencies and execution roles across domains, beyond adding a specialist teacher for each known task.

%% file: figures/overview.tex
\begin{figure}[t]
\centering
\includegraphics[width=0.52\textwidth]{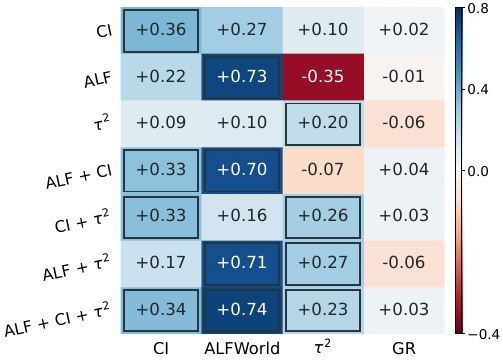}
\caption{Cross-domain accuracy changes relative to Base.
Rows denote student training sources; columns denote evaluation domains.
Outlined cells indicate domains included in training.
Absolute accuracies are reported in Appendix~\ref{app:results}.}
\label{fig:overview}
\end{figure}

%% file: tables/reasoning.tex
\begin{tabular}{lcccc}
\toprule
Model & Accuracy & \shortstack{Tool Call\\Rate} & \shortstack{Correct\\no tool} & \shortstack{Correct\\with tool} \\
\midrule
Base
    & $16.70$ & $70.40$ & $2.40$ & $14.30$ \\
ALF-OPD
    & $38.30$ & $11.50$ & $\tablebold{30.60}$ & $7.70$ \\
CI-OPD
    & $\tablebold{52.40}$ & $\tablebold{73.00}$ & $4.90$ & $\tablebold{47.60}$ \\
GR-OPD
    & $38.00$ & $14.40$ & $30.10$ & $7.90$ \\
\bottomrule
\end{tabular}

%% file: tables/positive_alf.tex
\begin{tabular}{lcccc}
\toprule
Model & Accuracy & Calls & Repeat & Change \\
\midrule
Base
    & $12.90$ & $12.97$ & $4.50$ & $90.80$ \\
CI
    & $\tablebold{40.20}$ & 18.94 & $\tablebold{0.70}$ & $\tablebold{94.30}$ \\
\bottomrule
\end{tabular}

%% file: tables/positive_tau.tex
\begin{tabular}{lcccccc}
\toprule
Model & All & Retail & Tool & Calls & Error & Query \\
\midrule
Base
    & $41.50$ & $48.10$ & $95.90$ & $5.66$ & $32.60$ & $54.50$ \\
CI
    & $\tablebold{51.00}$ & $\tablebold{65.30}$ & $95.80$ & $6.50$ & $\tablebold{23.50}$ & $\tablebold{67.50}$ \\
\bottomrule
\end{tabular}

%% file: tables/semantic_timing.tex
\begin{tabular}{lccc}
\toprule
Model & Accuracy & Ask--call & Dependency \\
\midrule
Base
    & $41.50$ & $\tablebold{0.04}$ & $0.50$ \\
Teacher
    & $46.00$ & $0.67$ & $0.38$ \\
ALF
    & $6.60$ & $94.30$ & $55.80$ \\
ALF+$\tau^2$
    & $\tablebold{68.90}$ & $0.24$ & $\tablebold{0.00}$ \\
\bottomrule
\end{tabular}

%% file: figures/dialogue.tex
\begin{tcolorbox}[title={ALF: generates a request but proceeds with a call},colback=red!2,colframe=black!65,fonttitle=\small\bfseries,boxrule=.4pt,left=4pt,right=4pt,top=3pt,bottom=3pt]
\small\textbf{Assistant:} \ldots\ Could you please provide your user ID?
\begin{lstlisting}[basicstyle=\ttfamily\footnotesize,breaklines=true,columns=fullflexible]
<tool_call>
{"name": "get_user_details", "arguments": {"user_id": "sara_doe_496"}}
</tool_call>
\end{lstlisting}
\textbf{Tool:} Error: User sara\_doe\_496 not found
\end{tcolorbox}
\begin{tcolorbox}[title={ALF+$\tau^2$: waits for the reply},colback=blue!2,colframe=black!65,fonttitle=\small\bfseries,boxrule=.4pt,left=4pt,right=4pt,top=3pt,bottom=3pt]
\small\textbf{Assistant:} \ldots\ Could you please provide your user ID?

\textbf{User:} Sure, my user ID is noah\_muller\_9847.

\textbf{Assistant:}
\begin{lstlisting}[basicstyle=\ttfamily\footnotesize,breaklines=true,columns=fullflexible]
<tool_call>
{"name": "get_user_details", "arguments": {"user_id": "noah_muller_9847"}}
</tool_call>
\end{lstlisting}
\end{tcolorbox}

%% file: tables/composition_scores.tex
\begin{tabular}{lcccc}
\toprule
Model & CI & ALF & $\tau^2$ & GR \\
\midrule
CI+GR
    & $\tablebold{51.30}$ & $38.60$ & $45.30$ & $44.50$ \\
ALF+CI
    & $49.80$ & $83.20$ & $34.00$ & $\tablebold{46.70}$ \\
CI+$\tau^2$
    & $50.20$ & $28.40$ & $67.80$ & $45.80$ \\
ALF+$\tau^2$
    & $33.80$ & $83.40$ & $\tablebold{68.90}$ & $36.70$ \\
ALF+CI+$\tau^2$
    & $51.10$ & $\tablebold{86.90}$ & $64.40$ & $45.00$ \\
\bottomrule
\end{tabular}

%% file: tables/composition_behavior.tex
\begin{tabular}{lccc}
\toprule
Model & $\tau^2$ score & Ask--call & Error repeat \\
\midrule
ALF+CI
    & $34.00$ & $25.60$ & $8.10$ \\
CI+$\tau^2$
    & $67.80$ & $0.66$ & $9.70$ \\
ALF+$\tau^2$
    & $\tablebold{68.90}$ & $0.24$ & $10.60$ \\
ALF+CI+$\tau^2$
    & $64.40$ & $\tablebold{0.10}$ & $\tablebold{3.00}$ \\
\bottomrule
\end{tabular}

%% file: sections_naacl/05_selecting_supervision.tex
\section{Exploring Mitigations for Negative Transfer in Agentic OPD}
\label{sec:supervision-exploration}
In this section, we mainly explore how to mitigate the potential cross-task negative transfer in agentic OPD training, using ALF-to-$\tau^2$ transfer (\S~\ref{sec:dependencies}) as the main case and extending the analysis to GR-to-CI transfer.

\subsection{Methodology: Selective OPD Supervision on Tokens with RL-Induced Preference}
\label{sec:preference-selection}
\label{sec:selection-results}
We summarize our core insight derived from the previous observations is agentic OPD training can teach student two parts of knowledge: (1) \textbf{\textit{generic agentic abilities}} (e.g., how to reason and act based on observations) and (2) \textbf{\textit{task-specific ReAct patterns}} (e.g., action habit of obtaining information in \S~\ref{sec:dependencies}). Severely overfitting to part (2) is the main cause of the negative transfer. Hence, the key idea of our proposed approach is to design a selective policy, which keeps the generic agentic abilities (mostly derived from the agentic RL training~\citep{yu2025demystifying}) and discards the plain transfer of surface-level task-specific ReAct patterns.

\begin{figure*}[bt]
 \centering
 \includegraphics[width=0.95\textwidth]{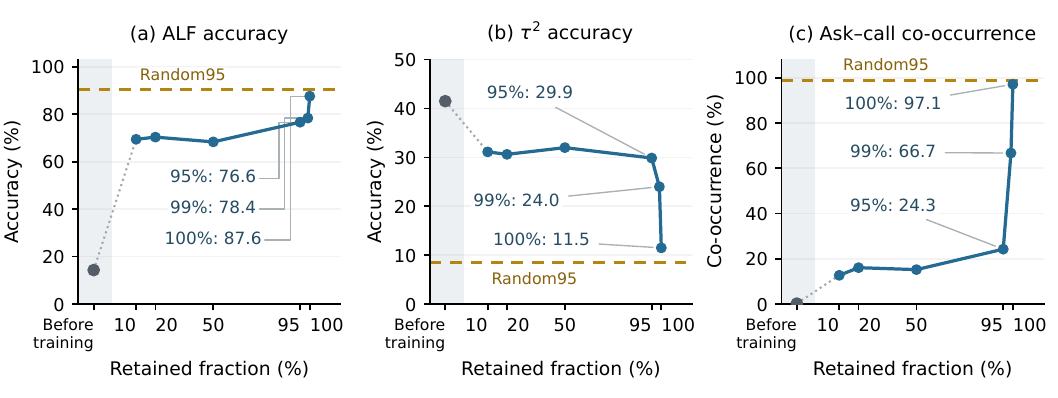} \caption{Effect of retained fraction after 20 updates on ALF accuracy, $\tau^2$ accuracy, and ask--call co-occurrence. The shaded leftmost point is an evaluation of the shared initialization from the Top95 run, before training. Trained points use their numerical retained fractions on the x-axis. Labels are offset for the closely spaced 95/99/100\% points. Horizontal dashed lines show the Random95 update-20 evaluation.}
 \label{fig:retained-fraction}
\end{figure*}

Given a student's rollout, we use both of the teacher's checkpoints before and after domain-specific RL to calculate log-likelihood on every token. Naturally, the difference of log-likelihood between two checkpoints (post-/pre-RL) indicates how agentic RL changes teacher models' preference on the target token. Let $\pi_{\mathrm{pre}}$ and $\pi_{\mathrm{post}}$ denote the two teacher checkpoints. For student-generated token $y_t$ under complete history $H_t$, define
\begin{equation*}
 \Delta_t=\log\frac{\pi_{\mathrm{post}}(y_t\mid H_t)}
                       {\pi_{\mathrm{pre}}(y_t\mid H_t)}.
 \label{eq:opd-decomposition}
\end{equation*}
Positive $\Delta$ indicates increased preference after RL; a strongly negative value indicates suppression. We rank positions from the highest to the lowest by signed $\Delta$ and gradually retain more high-rank tokens (i.e., initially retain top $10\%$, then top $20\%$, $30\%$, ...), while monitoring how student's performance change on ALFWorld and $\tau^2$-bench. Specifically, for a rollout with $N$ valid assistant positions $[0..T-1]$, retained fraction $\rho$ selects K$=\lceil\rho N\rceil$:
\begin{equation*}
 m_t=\mathbf1\!\left[t\in
 \operatorname{TopK}_{u\in[0..T-1]]}(\Delta_u)\right],
 \label{eq:delta-selection}
\end{equation*}
\begin{equation*}
 \mathcal L_{\mathrm{select}}(\theta)
 =\frac1K\sum_{0\leq t\leq T-1}m_t
 \ell_t^{\mathrm{OPD}}(\theta).
 \label{eq:selected-opd}
\end{equation*}
More detailed description and settings of our experiments can be found in Appendix~\ref{app:repro-selection}.
\paragraph{Results}
The results are shown in Figure~\ref{fig:retained-fraction}. We evaluate preference-guided selection using retained fractions of 10\%, 20\%, 50\%, 95\%, 99\%, and 100\%, where full retention corresponds to ordinary OPD. All runs use only ALFWorld training trajectories. We use $\tau^2$ evaluation to analyze the source--target trade-off across retention ratios. The key findings are concluded as following three phases: (\textbf{Phase 1}) By only retaining $10\%$ tokens, OPD can achieve about $80\%$ performance gain ($14.3\%\rightarrow 69.4\%$) on the training agentic task (i.e., ALFWorld). Meanwhile, the student model's $\tau^2$ performance ($41.5\%\rightarrow 31.1\%$) does not drop that drastically; (\textbf{Phase 2}) Gradually increasing the retained fraction ($10\%\rightarrow 95\%$), we observe that model's ALF accuracy slightly improves ($69.4\%\rightarrow 76.6\%$) while $\tau^2$ performance remains stable; (\textbf{Phase 3}) However, \textbf{\textit{retaining the final $5\%$ tokens (with least $5\%$ $\Delta$) starkly decreases}} the model's $\tau^2$ performance ($29.9\%\rightarrow 11.5\%$). Besides, we trace the variation of \textbf{\textit{ask--call co-occurence}} (as defined in \S~\ref{sec:dependencies}, measuring to what extent the models $\tau^2$ ReAct mode is interfered): Phase 3 has seen the steepest ascent of this metric ($24.3\%\rightarrow 97.1\%$). We also conduct a comparison experiment (“Random$95\%$”) by randomly retaining $95\%$ tokens and excluding the rest $5\%$ tokens. Dashed lines in Figure~\ref{fig:retained-fraction} show the results: unlike retaining $\Delta$ Top$95\%$ tokens, Random$95\%$ still affects student's $\tau^2$ performance.
\begin{tcolorbox}[title={Insight: Excluding Smallest $\Delta$ Tokens Can Rescue.},
colback=blue!2,colframe=blue!65,fonttitle=\small\bfseries,boxrule=.4pt]
\small Excluding a small fraction of tokens (i.e., with the least $5\%$ RL-induced $\Delta$) in the ALFWorld-domain OPD training can effectively mitigate the catastrophic negative transfer on the $\tau^2$ tasks, while maintaining most of the performance gain on ALFWorld tasks.
\end{tcolorbox}

\subsection{More Analysis and Explorations}
\label{sec:selection-analysis}
Section~\ref{sec:selection-results} shows that excluding a small fraction of positions helps alleviate large-scale performance degradation on $\tau^2$ bench. We present more analysis and explorations on this result. Detailed training dynamics and a separate GR prefix-supervision experiment are provided in Appendices~\ref{app:controls}.

\begin{wrapfigure}{r}{0.45\textwidth}
\centering
\includegraphics[width=0.86\linewidth]{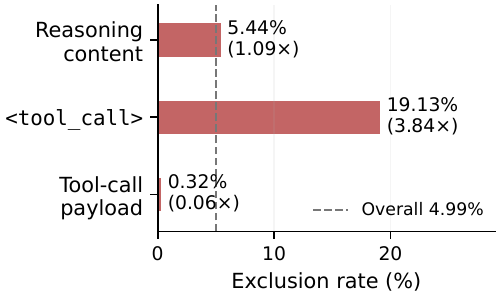} \caption{Exclusion rates and enrichment on Top95 training contexts. Reasoning content denotes assistant text outside tool-call blocks. \texttt{<tool\_call>} is the opening special token; payload denotes tokens inside a call.
Parentheses show enrichment relative to the overall exclusion rate (dashed line).}
\label{fig:delta-feedback}
\end{wrapfigure}

\paragraph{Which part of tokens are excluded?}
As shown in Figure~\ref{fig:delta-feedback}, reasoning content (text outside tool calls) tokens account for $98.19\%$ of excluded positions and $90.08\%$ of all valid positions, giving modest enrichment ($1.09\times$), while Tool-call onsets (the \texttt{<tool\_call>} special tokens) are more selectively affected: $19.13\%$ are excluded, or $3.84\times$ the overall exclusion rate. The specific tool-call content (tokens inside the \texttt{<tool\_call>} and \texttt{</tool\_call>}) are rarely excluded, with an exclusion rate of only 0.32\%. This pattern connects supervision to our ReAct observations. ALFWorld often obtains observations through environment actions, whereas $\tau^2$ tasks often require a user reply before acting. Suppressive updates to planning, state descriptions, and action onsets can change how the shared agent organizes its next step to obtain information.

\paragraph{Additional Explorations}
\label{sec:additional-explorations}
We first test three alternatives: \textbf{delimiter masking} removes supervision on the opening and closing tool-call tokens; \textbf{full-block masking} additionally removes the enclosed payload; and \textbf{negative masking} removes all negative ordinary OPD advantages while retaining the full-token denominator. These controls examine structural positions and feedback direction.

We further compare ranking rules using ordinary OPD feedback and the same selected-token normalization. \textbf{Advantage Top95} and \textbf{Advantage Top98} exclude the lowest $5\%$ and $2\%$ of signed ordinary advantages, respectively. The former matches the nominal exclusion fraction of $\Delta$ Top95. On fixed diagnostic contexts, the latter removes $43.56\%$ of absolute coefficient mass, compared with $41.02\%$ for $\Delta$ Top95; the retained mass after normalization is $57.54\%$ and $62.02\%$, respectively. This provides an approximately coefficient-comparable control, rather than a matched online gradient update. \textbf{Absolute-$\Delta$ Top95} instead excludes the $5\%$ of positions with the smallest $|\Delta|$, testing preference-change magnitude rather than direction.

\begin{table}[!htbp]
\caption{Results of alternative masking approaches.}
\label{tab:explorations}
\centering\small\setlength{\tabcolsep}{4pt}
\input{tables/supervision_explorations}
\end{table}

Advantage Top95 partially preserves $\tau^2$ accuracy ($23.50\%$), but ask--call co-occurrence remains high ($82.68\%$). Advantage Top98 and Absolute-$\Delta$ Top95 reach $5.88\%$ and $17.13\%$ $\tau^2$ accuracy, with co-occurrence of $99.16\%$ and $83.94\%$, respectively. All three achieve higher ALF accuracy ($81.66$--$85.86\%$) than $\Delta$ Top95 ($76.64\%$), while $\Delta$ Top95 retains higher $\tau^2$ accuracy ($29.88\%$) and substantially lower co-occurrence ($24.26\%$). Thus, neither filtering the strongest negative advantages nor selecting by change magnitude reproduces the same source--target balance in these comparisons. Appendix~\ref{app:ranking-controls-v22} provides definitions, coefficient diagnostics and learning curves for these three ranking controls.

\subsection{Extending Selection to GR-to-CI Transfer}
\label{sec:grci-selection-v22}
We next test whether selective supervision also changes the mathematical tool-use shift observed in Section~\ref{sec:dimensions}. For this extension, we train a Qwen3-30B-A3B teacher with tool-free GR reinforcement learning and use its pre-RL checkpoint as the reference. This teacher is distinct from the GR teacher in the earlier cross-domain matrix. Starting from the same 4B student initialization, Full OPD and signed-$\Delta$ Top95 use the same tool-free training data and are evaluated after 20 updates. The paired CI and GR evaluation sets contain the same mathematical questions, with a code interpreter available only in CI.

\begin{table}[!htbp]
\centering\small\setlength{\tabcolsep}{4pt}
\caption{Results of GR-to-CI transfer.}
\label{tab:grci-selection-v22}
\input{tables/grci_selection_v22}
\end{table}

Table~\ref{tab:grci-selection-v22} shows similar GR accuracy ($46.02\%$ versus $46.33\%$), while Top95 increases CI tool attempts from $10.41\%$ to $29.49\%$. The proportion of CI samples that use a tool and answer correctly increases from $6.02\%$ to $11.63\%$. Overall CI accuracy is lower, however ($36.53\%$ versus $37.65\%$), with fewer correct answers on the no-tool route ($24.90\%$ versus $31.63\%$). Selection therefore preserves more tool-assisted behavior without improving aggregate CI accuracy in this setting. Together with ALF-to-$\tau^2$, this extension shows why task accuracy and interaction behavior should be assessed jointly. Appendix~\ref{app:grci-selection-v22} reports the recipe and learning curves.

%% file: tables/supervision_explorations.tex
\begin{tabular}{lccc}
\toprule
Approach & ALF acc. & $\tau^2$ acc. & Ask--call rate \\
\midrule
Signed-$\Delta$ Top95
    & $76.64$ & $\mathbf{29.88}$ & $\mathbf{24.26}$ \\
Delimiter masking
    & $76.28$ & $4.88$ & $99.37$ \\
Full-block masking
    & $79.74$ & $3.00$ & $99.72$ \\
Negative masking
    & $80.02$ & $3.63$ & $83.54$ \\
\midrule
Advantage Top95
    & $81.66$ & $23.50$ & $82.68$ \\
Advantage Top98
    & $83.03$ & $5.88$ & $99.16$ \\
Absolute-$\Delta$ Top95
    & $\mathbf{85.86}$ & $17.13$ & $83.94$ \\
\bottomrule
\end{tabular}

%% file: tables/grci_selection_v22.tex
\begin{tabular}{lccccc}
\toprule
Method & GR acc. & CI acc. & CI tool attempts & No tool \& correct & Tool \& correct \\
\midrule
Full OPD
    & $46.02$ & $\textbf{37.65}$ & $10.41$ & $\textbf{31.63}$ & $6.02$ \\
Signed-$\Delta$ Top95
    & $\textbf{46.33}$ & $36.53$ & $\textbf{29.49}$ & $24.90$ & $\textbf{11.63}$ \\
\bottomrule
\end{tabular}

%% file: sections_naacl/07_react_conclusion.tex
\section{Conclusion}

Agentic OPD reshapes how LLM agents reason, act, and obtain observations across domains, producing positive and negative transfer. Task-matched MOPD mitigates the conversational failures induced by ALF supervision on $\tau^2$. As a source-only training exploration, we select ordinary OPD supervision positions using pre/post-domain-RL preference changes. Excluding the low-$\Delta$ tail substantially alleviates negative transfer while sacrificing source-domain performance. Token-level analysis reveals concentrated context-dependent suppressive feedback, including reasoning and tool-call onsets. Structural masks, negative masking, and alternative ranking rules yield different source–target trade-offs; none of the reported alternatives reproduces the same target-interaction preservation as signed-Δ Top95 in the ALF comparison. An extension with a RL-trained GR teacher preserves more CI tool use at equal mean GR accuracy. These findings highlight the importance of controlling where specialization feedback is applied in agentic OPD.

%% file: appendix_new/discussion.tex
\section{Discussion}
\label{sec:discussion}

\paragraph{What should agentic training evaluate?}
Task accuracy, action content, and feedback dependencies capture different properties of agent behavior. The mathematical results pair improved accuracy with fewer tool attempts; CI transfer pairs better outcomes with action adjustment and query coverage. Conversely, ALF transfer produces request-like text alongside tool calls, some of which depend on user information or authorization not yet obtained. Evaluation should therefore measure how an agent prepares an action, acquires its prerequisites, and responds to the resulting observation, alongside task utility \citep{arabzadeh2024assessing} and benchmark breadth \citep{liu2024agentbench,zhou2024webarena}. Importantly, ask--call co-occurrence captures generated output patterns rather than actual user requests, and should be interpreted alongside dependency violations and task accuracy.

\paragraph{How should supervision be composed?}
MOPD provides complementary training coverage: CI supports mathematical tool use and feedback-driven actions, while task-matched $\tau^2$ supervision mitigates failures to wait for required user information. These findings connect teacher composition to behavioral coverage and task grouping \citep{yu2020gradient,standley2020tasks}. Source-only $\Delta$-guided selection offers another approach when target-domain training data are unavailable. Excluding a small low-$\Delta$ tail reduces concentrated suppressive feedback at context-dependent positions, with a source-learning trade-off. Delimiter masking, full-block masking, and negative masking fail to reproduce the same source--target balance. These results highlight the importance of feedback location and direction, while further controlled interventions are needed to isolate the mechanisms underlying cross-domain transfer.

%% file: appendix_new/experimental_setup.tex
\section{Experimental Setup and Training}
\label{app:repro-training}
\label{app:training}

Our main experimental settings are concluded in Table~\ref{tab:settings}.

\begin{table}[!ht]
\centering
\small \caption{ Interaction approaches in the four training domains. Their feedback sources and turn requirements are described below. }
\label{tab:settings}
\setlength{\tabcolsep}{4pt}
\begin{tabular}{
    @{}p{0.10\columnwidth}
    p{0.82\columnwidth}@{}
}
\toprule
Domain & Interaction approach \\
\midrule
GR &
Text-based mathematical reasoning without an external
tool-feedback loop. \\[3pt]
CI &
Reasoning interleaved with Python execution for
calculation, symbolic manipulation, and verification. \\[3pt]
ALF &
Reasoning followed by an admissible environment action,
which returns the next observation. \\[3pt]
$\tau^2$ &
Customer communication and account-side tool use,
with actions conditioned on available information
and required authorization. \\
\bottomrule
\end{tabular}
\end{table}

\subsection{Domain Interaction and Tool-Use Instructions}
\label{app:tool-prompts}

The domains differ in how reasoning leads to an action and how the agent obtains the feedback needed for its next decision.

\paragraph{General reasoning (GR).}
GR training uses mathematical problems solved through natural-language reasoning. The model develops a solution and produces its answer without an external tool-feedback loop. This source therefore supervises text-based derivation rather than interaction with an execution environment.

\paragraph{Code interpretation (CI).}
CI allows the agent to invoke \texttt{code\_interpreter} when calculation, symbolic manipulation, or verification is useful. A tool call supplies Python code, and its execution result is appended to the interaction history. The agent can then use that result to continue reasoning, revise a calculation, or produce its answer.

\paragraph{ALFWorld (ALF).}
At each decision step, the agent receives an environment observation and a list of admissible actions. The prompt asks it to reason about the current situation and execute exactly one listed action through \texttt{alfworld\_step}. The action argument must match an entry in the admissible action list. Execution returns the next observation, so information gathering and task progress are both mediated by environment actions.

\paragraph{Conversational tool use ($\tau^2$).}
The agent alternates between communicating with a customer and invoking an available account-side tool. The prompt specifies one interaction branch per turn: a customer-facing message or a tool call. Missing information must be obtained from the customer, and database-changing actions require explicit confirmation under the corresponding domain policy. Customer-side device procedures are communicated as instructions for the customer to perform. Airline, retail, and telecom share these interaction requirements while using different policies and tools.

Tool calls in CI, ALF, and $\tau^2$ use JSON objects with \texttt{name} and \texttt{arguments} fields enclosed by \texttt{<tool\_call>} and \texttt{</tool\_call>} markers. This serialization is shared across the tool-using domains; the surrounding reasoning, feedback source, and turn requirements depend on the task.

\subsection{Model Initialization and Teacher Construction}
\label{app:repro-teachers}

\begin{figure*}[t]
\centering
\includegraphics[width=0.94\textwidth]{
    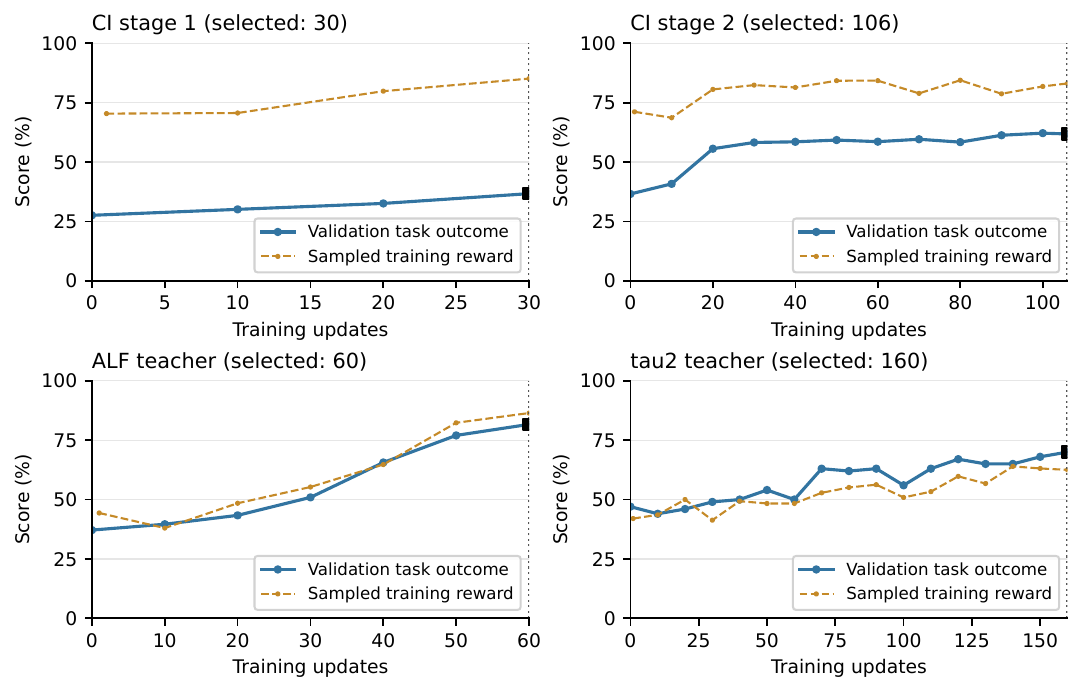
}
\caption{ Teacher RL learning curves. Solid curves show validation task outcomes, and dashed curves show sampled training reward. CI validation measures answer accuracy, with separate update axes for its two training stages. Squares mark the selected teacher checkpoints. Training and validation use their respective prompt pools and decoding settings. }
\label{fig:repro-teacher-curves}
\end{figure*}

All students initialize from Qwen3-4B-Instruct-2507. The CI teacher is a domain-specialized model from the same 4B family. The ALF and $\tau^2$ teachers initialize from Qwen3-30B-A3B-Instruct-2507, with approximately 30B total and 3B active parameters, and undergo domain-specific agentic RL. GR uses the released general 30B-A3B model. Teachers remain frozen during OPD. Figure~\ref{fig:repro-teacher-curves} shows teacher learning during specialization and marks the checkpoints used for distillation. Tables~\ref{tab:repro-optimizer} summarize the optimization settings.

For preference-guided supervision, the ALF teacher's checkpoint immediately before domain-specific RL serves as the pre-domain-RL reference, and its specialized checkpoint serves as the post-domain-RL teacher.

\paragraph{CI teacher.}
CI specialization uses two GRPO stages on a 13,616-row English subset of DAPO-Math-17k\citep{yu2026dapo} with Python execution. Stage 1 combines answer verification with tool-use shaping and a repetition penalty. Stage 2 initializes from the stage-1 weights at update 30 with a fresh optimizer and retains answer verification and the repetition penalty. The selected CI teacher is the stage-2 checkpoint at update 106.

Let $r_{\rm ans}$ be the answer-verifier reward, $n_{\rm clean}$ the number of tool responses passing the response-error detector, and $I$ an indicator. The stage rewards are
\begin{align}
R_1 ={}&
    r_{\rm ans}
    -0.30 I_{\rm repeat}
    +0.05\min(n_{\rm clean},3)
    \nonumber\\
    &-0.10 I_{\rm fail}
    +0.05 I_{\rm pre}
    +0.05 I_{\rm early},\\
R_2 ={}&
    r_{\rm ans}-0.30 I_{\rm repeat}.
\end{align}
The repetition detector checks whether the final 512-character block appears earlier in an output of at least 1,024 characters. $I_{\rm fail}$ indicates that tools are called but no response passes the error detector. $I_{\rm pre}$ requires at least 500 characters before the first call. $I_{\rm early}$ requires a clean execution before the first boxed answer, or a clean execution when no boxed answer is produced.

\paragraph{ALF and $\tau^2$ teachers.}
ALF specialization uses 3,553 prompts and binary environment task-completion reward. The $\tau^2$ training collection contains 2,160 rows: 1,178 airline, 637 retail, and 345 telecom. Its reward follows the benchmark's task checks, including database and communication requirements where applicable. The selected teachers are the ALF checkpoint at update 60 and the $\tau^2$ checkpoint at update 160.

\subsection{On-Policy Distillation Implementation}
\label{app:repro-opd}
\label{app:repro-implementation}

The student generates trajectories using the interaction approach of each training domain. A frozen teacher scores the student-generated assistant tokens under the same complete histories, including previous assistant outputs, tool observations, and user replies. In MOPD, each prompt is scored by its assigned source-domain teacher.

Let $\pi_T$ denote that teacher and $\pi_{\theta_{\rm old}}$ the rollout student. The ordinary token advantage is
\begin{equation}
    A_{T,t}
    =
    \log\pi_T(y_t\mid H_t)
    -
    \log\pi_{\theta_{\rm old}}(y_t\mid H_t).
    \label{eq:opd-advantage}
\end{equation}
The implementation clips this advantage and detaches the resulting coefficient $\bar A_t$. For
$r_t=\pi_\theta(y_t\mid H_t)/
\pi_{\theta_{\rm old}}(y_t\mid H_t)$,
the clipped surrogate is
\begin{equation}
    b_t=
    \max\left\{
        -r_t\bar A_t,\,
        -\operatorname{clip}
        (r_t,1-\epsilon_l,1+\epsilon_u)\bar A_t
    \right\}.
\end{equation}
Dual clipping gives
\begin{equation}
    \ell_t^{\mathrm{OPD}}=
    \begin{cases}
        \min(b_t,-\beta\bar A_t),
            & \bar A_t<0,\\
        b_t,
            & \bar A_t\geq0.
    \end{cases}
\end{equation}
The advantage clamp, ratio bounds, and dual-clip coefficient are listed in Table~\ref{tab:repro-optimizer}. The log policy ratio is bounded to $[-20,20]$ before exponentiation.

Let $q_{jt}$ identify assistant positions with teacher scores that are retained for supervision. The loss averages retained-token means across rollouts:
\begin{equation}
    \mathcal L
    =
    \frac{1}{B}
    \sum_{j=1}^{B}
    \frac{
        \sum_t q_{jt}\ell_{jt}^{\mathrm{OPD}}
    }{
        \sum_t q_{jt}+10^{-8}
    }.
    \label{eq:opd-masked-loss}
\end{equation}
Tool observations, user replies, and padding remain available as context but receive no distillation loss. Sequences with no retained positions contribute zero. Task outcomes are recorded for monitoring; the update uses the distillation advantage.

\input{tables/optimizer_recipe.tex}
\input{tables/opd_recipe.tex}

\paragraph{Source mixtures.}
Pairwise CI+ALF and CI+GR training use 3,000 prompts from each source. CI+$\tau^2$ and ALF+$\tau^2$ combine 3,000 prompts from the first source with 2,160 $\tau^2$ prompts. Three-teacher training combines 3,000 CI, 3,000 ALF, and 2,160 $\tau^2$ prompts. The collections are concatenated and shuffled, with teacher routing determined by each prompt's source. These counts determine prompt-level sampling proportions; trajectory lengths determine the corresponding token exposure.

\subsection{Preference-Guided Supervision}
\label{app:repro-selection}

Preference-guided experiments use only the 3,000-prompt ALF training subset. The frozen pre-domain-RL reference and post-domain-RL teacher score the same student-generated positions.

For each multi-turn rollout, selection ranks valid assistant positions across all turns by signed $\Delta$, as defined in Section~\ref{sec:preference-selection}. A retained fraction $\rho$ selects $K=\lceil\rho N\rceil$ of the $N$ valid positions. Ties prioritize the earlier response position. Retained positions receive the ordinary OPD advantage, including both positive and negative feedback, and their loss is normalized by the selected count.

Random95 draws a uniform subset without replacement using the same retained count and normalization. Top100 retains all valid positions. The remaining training settings are given in Table~\ref{tab:repro-opd-recipe}. Definitions of the structural and feedback controls appear in Appendix~\ref{app:controls}.

\subsection{Checkpoint Selection and Learning Dynamics}
\label{app:repro-checkpoints}

\begin{figure*}[t]
\centering
\includegraphics[width=0.94\textwidth]{
    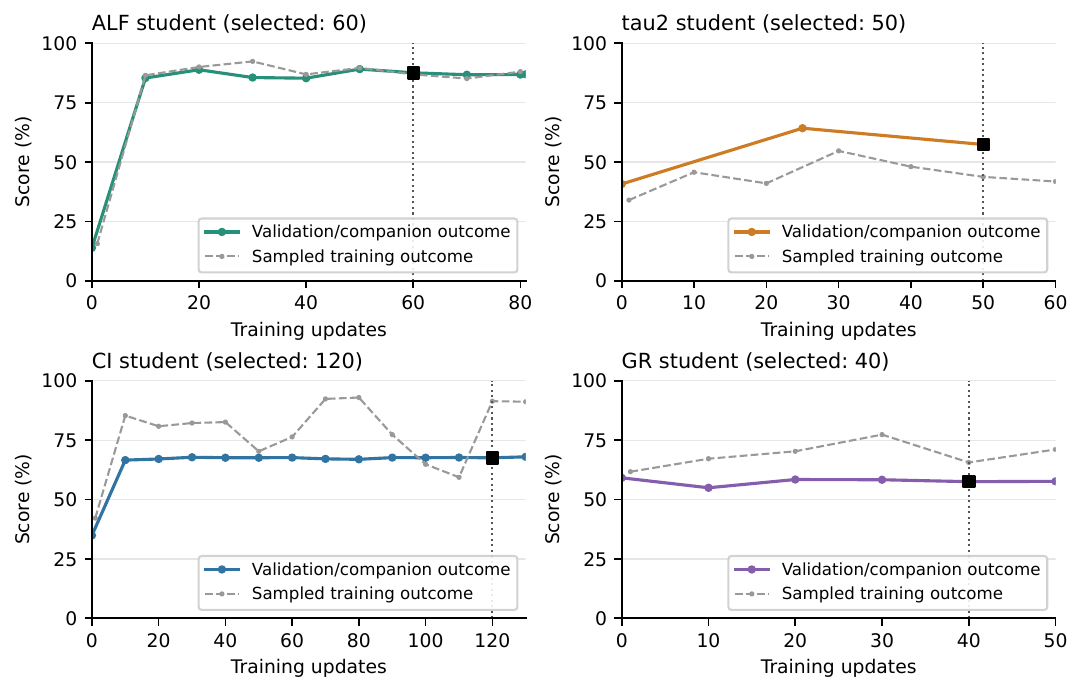
}
\caption{ Single-teacher OPD learning curves. Solid curves show source-validation outcomes, and dashed curves show sampled training outcomes. Squares mark the checkpoints used in the main analyses. CI is displayed through update 130 and GR through update 50. The $\tau^2$ validation curve contains measurements at updates 0, 25, and 50. Mathematical monitoring pools differ from the 980-trajectory behavioral evaluation subset. }
\label{fig:repro-opd-single}
\end{figure*}

\begin{figure*}[t]
\centering
\includegraphics[width=0.94\textwidth]{
    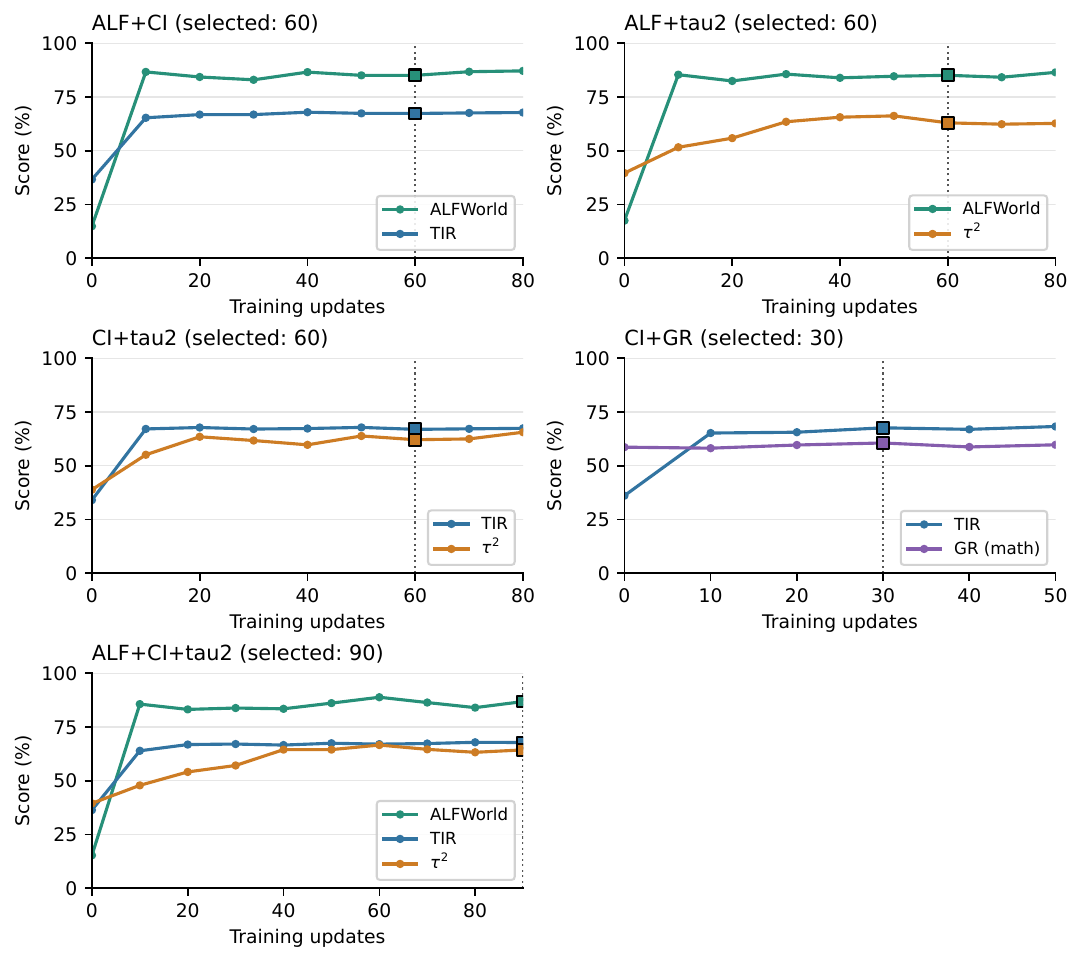
}
\caption{ Source-validation learning curves for teacher compositions. Each curve reports a separate training domain; squares mark the checkpoints used in the main analyses. ALF+CI and CI+$\tau^2$ are displayed through update 80, and CI+GR through update 50. Other panels show their respective recorded ranges. }
\label{fig:repro-opd-mixtures}
\end{figure*}

Each named student uses a fixed representative checkpoint selected from its validation stage. The same checkpoint is used for its task outcomes and behavioral analyses. Figures~\ref{fig:repro-opd-single} and~\ref{fig:repro-opd-mixtures} show the learning curves and mark these checkpoints.

The curves document source learning for single-teacher and multi-teacher training. For compositions, each domain is monitored separately, so progress in one source is visible alongside progress in the others. Preference-guided comparisons instead use the common update-20 checkpoint specified in Section~\ref{sec:preference-selection}. Evaluation datasets, decoding settings, and behavioral metrics are described in Appendix~\ref{app:evaluation}.

%% file: tables/optimizer_recipe.tex
\begin{table*}[t]
\caption{Optimization settings separated from sampling and task
rewards. Adam momentum and epsilon are inherited by the backend
when no explicit optimizer override is provided. Configured
minibatch sizes and rollout counts are listed separately because
the worker expands prompt-unit minibatches by the rollout count.}
\label{tab:repro-optimizer}
\centering\small\setlength{\tabcolsep}{5pt}
\begin{tabular}{p{0.35\textwidth}p{0.26\textwidth}p{0.27\textwidth}}
\toprule
Parameter & Teacher RL & OPD / selected-position runs\\
\midrule
Optimizer & Megatron Adam & Megatron Adam\\
Learning rate & $10^{-6}$ & $10^{-5}$\\
Warmup / schedule & none / constant & none / constant\\
Weight decay & 0.01 & 0.01\\
Adam momentum & backend $(0.9,0.999)$ & backend $(0.9,0.999)$\\
Adam epsilon & backend default; no override & backend default; no override\\
Gradient-norm clip & 1.0 & 1.0\\
Policy-ratio clip / dual clip & $[0.8,1.28]$ / 10 & $[0.8,1.28]$ / 10\\
OPD advantage clamp & not applicable & $[-10,10]$\\
Entropy coefficient & 0 & 0\\
Explicit KL loss / KL in reward & both disabled & both disabled\\
Task-reward mixing & GRPO reward advantage & 0 (monitoring only)\\
Distillation coefficient & not applicable & 1\\
Precision & bfloat16 & bfloat16\\
Dynamic token microbatching & enabled & enabled\\
Teacher gradients & trainable RL policy & frozen scoring only\\
\bottomrule
\end{tabular}
\end{table*}

%% file: tables/opd_recipe.tex
\begin{table*}[t]
\caption{Main OPD and selected-position launch recipes. Batch and
configured PPO minibatch are prompt counts; ``Samples'' is rollouts
per prompt. Learning rate is $10^{-5}$ and per-GPU microbatch is one.
Prompt/response caps are measured in tokens.}
\label{tab:repro-opd-recipe}
\centering\small\setlength{\tabcolsep}{3pt}
\begin{tabular}{lrrrrr}
\toprule
Student/condition & Rows & Batch & Samples & Traj./batch & Prompt/response caps\\
\midrule
CI & 17,917 & 128 & 4 & 512 & 1,024/31,740\\
ALF & 3,553 & 128 & 4 & 512 & 1,024/31,740\\
$\tau^2$ & 2,160 & 32 & 8 & 256 & 8,192/32,768\\
GR & 17,917 & 32 & 4 & 128 & 8,192/32,768\\
CI+GR & 6,000 & 128 & 4 & 512 & 8,192/32,768\\
ALF+CI & 6,000 & 128 & 4 & 512 & 1,024/31,740\\
CI+$\tau^2$ & 5,160 & 128 & 4 & 512 & 8,192/32,768\\
ALF+$\tau^2$ & 5,160 & 128 & 4 & 512 & 8,192/32,768\\
ALF+CI+$\tau^2$ & 8,160 & 128 & 4 & 512 & 8,192/32,768\\
Top95/Random95/Top100 & 3,000 & 32 & 4 & 128 & 8,192/32,768\\
\bottomrule
\end{tabular}
\end{table*}

%% file: appendix_new/evaluation_protocols.tex
\section{Evaluation Protocols and Behavioral Metrics}
\label{app:evaluation}
\label{app:repro-evaluation}

\subsection{Evaluation Data and Execution}

Mathematical evaluation contains 30 AIME2026~\citep{aime26} questions sampled 16 times and 100 BeyondAIME~\citep{bytedance_seed_2025_beyondaime} questions sampled five times, giving 980 trajectories per model. The tool-enabled setting provides \texttt{code\_interpreter}; the no-tool setting evaluates text-based solutions. ALFWorld evaluation contains 140 in-distribution and 134 out-of-distribution tasks, each evaluated four times, yielding 1,096 trajectories per model. The $\tau^2$ evaluation contains 100 tasks sampled eight times: 160 airline, 320 retail, and 320 telecom trajectories. We use the same sampling settings for all training rollouts and evaluations: temperature $1.0$, top-$p$ $1.0$, and unrestricted top-$k$. Teacher token scoring computes probabilities for student-generated tokens at temperature $1.0$, without generating additional responses.

Agent execution uses the Qwen3 chat template and structured tool calls. The loop parses the complete assistant output before choosing the next interaction branch. In $\tau^2$, exactly one parsed call with object-valued arguments is submitted to the environment. If prose accompanies the call, it remains in the agent history but is not sent to the customer. Text-only outputs are sent to the customer. Multiple parsed calls, or a tool marker with no parseable call, produce local error feedback.

The $\tau^2$ user simulator uses DeepSeek-V4-Flash~\citep{xu2026deepseek} at temperature 0.7. Natural-language assertions use GPT-5.5 at temperature 1.0.

\subsection{Metric Definitions}

Table~\ref{tab:behavioral-metrics} defines the outcome, generation, and interaction metrics. Their denominators follow the event being measured: task outcomes use trajectories, request--call co-occurrence uses request turns, and feedback-response metrics use eligible action transitions.

\begin{table*}[t]
\centering
\caption{ Definitions of the evaluation metrics. Calls are reported as a mean count; rates are reported as percentages. Generated calls and JSON validity describe model outputs, while task accuracy measures final outcomes. }
\label{tab:behavioral-metrics}
\small \setlength{\tabcolsep}{5pt}
\renewcommand{\arraystretch}{1.15}
\begin{tabular}{
    @{}p{0.19\textwidth}
    p{0.28\textwidth}
    p{0.46\textwidth}@{}
}
\toprule
Metric & Unit or denominator & Definition \\
\midrule

Task accuracy &
Evaluation trajectories &
Mean answer correctness for mathematics or binary
task completion for ALFWorld and $\tau^2$. \\

Tool-attempt rate &
Evaluation trajectories &
Fraction containing at least one generated
tool-call marker. \\

Calls per trajectory &
Evaluation trajectories &
Mean number of generated spans enclosed by opening
and closing tool-call markers. \\

Call JSON validity &
Complete generated call blocks &
Whether a block parses to an object with a
string-valued name and dictionary-valued arguments. \\

Ask--call co-occurrence &
Assistant outputs containing detected request-like text &
Fraction of assistant outputs containing detected request-like text that also contain a generated tool call. \\

Dependency-violation rate &
Evaluation trajectories &
Fraction of trajectories containing a tool call that depends on user information or authorization not yet obtained, as identified through semantic review. \\

Affected-task coverage &
Benchmark tasks &
Fraction with at least one sampled trajectory
containing a dependency violation. \\

Adjacent-action repetition &
Consecutive ALF action pairs &
Fraction with identical normalized action strings. \\

Post-no-op adjustment &
ALF no-op observations followed by another action &
Fraction whose next action differs from the action
that produced the no-op observation. \\

Post-error repetition &
Tool-error observations followed by an assistant turn &
Fraction whose next parsed call repeats the preceding
call's tool name and normalized arguments. \\

Error-bearing trajectory rate &
Evaluation trajectories &
Fraction containing at least one tool-error
observation. \\

Reference-query coverage &
Eligible retail trajectories ($n=200$) &
Fraction containing every annotated reference query,
matched by tool name and normalized arguments. \\

\bottomrule
\end{tabular}
\end{table*}

\paragraph{Mathematical solution routes.}
Correct outputs are partitioned by the presence of a tool attempt. Both shares use all 980 trajectories as their denominator, so correct tool-containing and correct no-tool outputs sum to overall accuracy.

\paragraph{Information requests and dependencies.}
The request detector identifies request-like expressions in generated assistant text after removing complete tool-call blocks. These include requests to provide, confirm, share, or supply information, as well as expressions such as ``I need your'' followed by an identifier or phone number. Ask--call co-occurrence measures the fraction of outputs containing detected request-like text that also contain a generated tool call. Importantly, in our execution protocol, text accompanying a tool call is not delivered to the user. Thus, this metric captures the co-occurrence of request-like text and tool calls in model outputs, rather than requests actually communicated to users. The dependency-violation metric further identifies calls that require information or authorization not yet obtained from the user.

\paragraph{Responses to feedback.}
ALF action comparisons use normalized action strings. For post-error repetition, eligible feedback contains \texttt{Error:}; the first parsed call in the following assistant turn is compared with the last parsed call in the preceding turn. These metrics measure how the agent changes its actions after an environment response.

\paragraph{Reference-query coverage.}
The retail subset contains trajectories with annotated \texttt{get\_*} or \texttt{find\_*} reference actions. Coverage requires matching all reference queries somewhere in the generated calls. It captures whether the agent retrieves the required information, independently of query order.

\subsection{Semantic Review of Dependency Violations}
\label{app:semantic}

We review 800 natural $\tau^2$ trajectories each from Base, the ALF teacher, the ALF student, and the ALF+$\tau^2$ student. Qwen3.8-Max assesses tool-call attempts using the domain policy, tool schemas, preceding history, and current assistant output. Model identity, final task outcome, and future feedback are withheld. An independent human annotator additionally reviews 120 stratified tool-call events, including predicted violations, non-violations, and ambiguous cases, without access to model identities or task outcomes. Human judgments agree with the LLM reviewer on 118 of the 120 examined events.

A current-request dependency violation occurs when a call requires the information or authorization requested in that same turn. Values appearing only in schema examples or assistant assertions do not establish that the user has supplied them. Calls independent of the current request are assessed separately. Table~\ref{tab:semantic-counts} reports the turn-level counts and their coverage across trajectories and tasks.

\begin{table*}[t]
\centering
\caption{ Natural-trajectory semantic review. The first two columns report turn-level numerator/denominator pairs. Rollout coverage counts trajectories containing at least one dependency violation. Task coverage counts tasks with at least one affected trajectory among their eight samples. }
\label{tab:semantic-counts}
\label{tab:dependency-coverage}
\small \setlength{\tabcolsep}{6pt}
\begin{tabular}{lrrrr}
\toprule
Model &
\shortstack{Ask+call\\ / Ask} &
\shortstack{Dependent\\ / Ask+call} &
\shortstack{Affected rollouts\\ / 800} &
\shortstack{Affected tasks\\ / 100} \\
\midrule
Base
    & 1/2,278
    & 0/1
    & 4
    & 4 \\
ALF teacher
    & 14/2,081
    & 4/14
    & 3
    & 3 \\
ALF
    & 5,022/5,328
    & 4,411/5,022
    & 446
    & 98 \\
ALF+$\tau^2$
    & 5/2,055
    & 0/5
    & 0
    & 0 \\
\bottomrule
\end{tabular}
\end{table*}

For ALF, 87.8\% of detected ask--call turns contain a call dependent on the unanswered request. Violations affect 98 of the 100 tasks, showing that the pattern extends across the evaluation set. The ALF+$\tau^2$ student has no detected current-request dependency violations in these reviewed trajectories. Semantic review also covers attempts outside the lexical request detector. This accounts for Base's four affected trajectories despite no violation in its single detected ask--call turn.

%% file: appendix_new/evaluation_results.tex
\section{Complete Evaluation Results}
\label{app:results}

Table~\ref{tab:full-results} gives the absolute accuracies underlying the cross-domain heatmap and teacher-composition analysis. Each named student uses the checkpoint specified in Appendix~\ref{app:repro-checkpoints}. Table~\ref{tab:tau-domains} separates $\tau^2$ performance into airline, retail, and telecom.

\begin{table}[t]
\centering
\caption{ Complete accuracies for the selected students. }
\label{tab:full-results}
\small \setlength{\tabcolsep}{2.5pt}
\begin{tabular}{lrrrr}
\toprule
Model &
\shortstack{CI\\$n=980$} &
\shortstack{ALF\\$n=1{,}096$} &
\shortstack{$\tau^2$\\$n=800$} &
\shortstack{GR\\$n=980$} \\
\midrule
Base
    & 0.167 & 0.129 & 0.415 & 0.424 \\
GR
    & 0.380 & 0.214 & 0.425 & 0.413 \\
CI+GR
    & 0.513 & 0.386 & 0.453 & 0.445 \\
CI
    & 0.524 & 0.402 & 0.510 & 0.444 \\
ALF
    & 0.383 & 0.855 & 0.066 & 0.417 \\
$\tau^2$
    & 0.262 & 0.229 & 0.611 & 0.367 \\
ALF+CI
    & 0.498 & 0.832 & 0.340 & 0.467 \\
CI+$\tau^2$
    & 0.502 & 0.284 & 0.678 & 0.458 \\
ALF+$\tau^2$
    & 0.338 & 0.834 & 0.689 & 0.367 \\
ALF+CI+$\tau^2$
    & 0.511 & 0.869 & 0.644 & 0.450 \\
\bottomrule
\end{tabular}
\end{table}

\begin{table}[t]
\centering
\caption{ Per-domain $\tau^2$ task accuracy. Teacher rows provide reference performance; the remaining rows are 4B students. }
\label{tab:tau-domains}
\small \setlength{\tabcolsep}{4pt}
\begin{tabular}{lrrr}
\toprule
Model &
\shortstack{Airline\\$n=160$} &
\shortstack{Retail\\$n=320$} &
\shortstack{Telecom\\$n=320$} \\
\midrule
Base
    & 0.325 & 0.481 & 0.394 \\
ALF teacher
    & 0.319 & 0.634 & 0.356 \\
$\tau^2$ teacher
    & 0.469 & 0.684 & 0.822 \\
CI
    & 0.425 & 0.653 & 0.409 \\
ALF
    & 0.019 & 0.119 & 0.037 \\
$\tau^2$
    & 0.369 & 0.528 & 0.816 \\
ALF+CI
    & 0.125 & 0.400 & 0.388 \\
CI+$\tau^2$
    & 0.463 & 0.672 & 0.791 \\
ALF+$\tau^2$
    & 0.369 & 0.678 & 0.859 \\
ALF+CI+$\tau^2$
    & 0.275 & 0.662 & 0.809 \\
\bottomrule
\end{tabular}
\end{table}

CI's largest $\tau^2$ gain occurs in retail. The ALF teacher also performs substantially better than the ALF-trained student on $\tau^2$, consistent with the distinction between teacher competence and its transfer through source-domain supervision.

%% file: appendix_new/analysis_preference_guided_selection.tex
\section{Token-Level Analysis of Preference-Guided Selection}
\label{app:selection-analysis}

We analyze a batch of rollouts sampled during $\Delta$ Top95 training. All comparisons below use the same token sequences and model scores. For upper-tail comparisons, each rollout contributes as many highest-$\Delta$ positions as excluded positions.

\subsection{Feedback Concentration and Direction}
\label{app:selection-coefficients}

We measure feedback using the clipped ordinary advantage at unit policy ratio:
\begin{equation}
    c_t
    =
    -\frac{
        \operatorname{clip}(A_{\mathrm{post},t},-10,10)
    }{N},
\end{equation}
where $N$ is the number of valid assistant positions. The excluded share of absolute coefficient mass is
\begin{equation}
    R_{\mathrm{excluded}}
    =
    \frac{
        \sum_t(1-m_t)|c_t|
    }{
        \sum_t|c_t|
    }.
\end{equation}
We average this ratio equally across rollouts. The statistic measures frozen feedback coefficients before importance weighting and optimizer transformations.

Top95 excludes 41.02\% of ordinary absolute coefficient mass, compared with 5.56\% for Random95. Of the mass excluded by Top95, 39.59 percentage points come from negative advantages and 1.43 from positive advantages. Selection therefore primarily removes feedback that suppresses student-generated continuations.

Negative advantages account for 88.49\% of ordinary mass on these contexts. Top95 retains 48.90 percentage points as negative feedback, preserving correction at other positions. After selected-token normalization, retained absolute mass is 62.02\% of ordinary OPD for Top95 and 99.33\% for Random95.

\begin{figure}[t]
\centering
\includegraphics[width=0.55\textwidth]{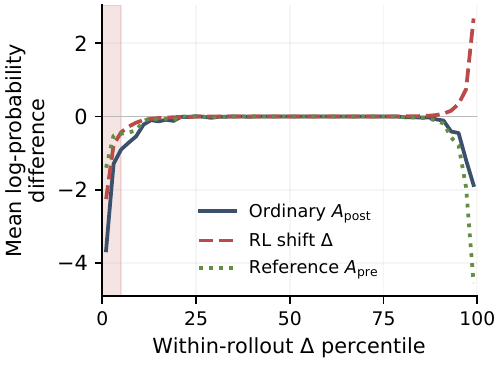} \caption{RL-induced log-probability changes and OPD advantages on Top95 training contexts. Curves show mean ordinary advantage, pre-domain-RL reference advantage, and signed $\Delta$ across within-rollout preference percentiles. Shading marks the excluded tail.}
\label{fig:delta-feedback-advantages}
\end{figure}

\paragraph{RL-induced preference changes and OPD feedback.}
Figure~\ref{fig:delta-feedback-advantages} relates the selection signal to ordinary OPD feedback. Positions are ranked by signed $\Delta$ within each rollout, and the curves average the corresponding bin means equally across rollouts.

In the lowest-$\Delta$ tail, both the pre-domain-RL reference and the specialist assign negative advantages on average, with the specialist providing stronger suppression. Domain-specific RL therefore strengthens existing corrective feedback at many excluded positions. For 16.05\% of excluded positions, it also reverses the feedback direction: $A_{\mathrm{pre},t}>0$ while $A_{\mathrm{post},t}<0$. Together with the coefficient-mass analysis above, this pattern shows that the excluded tail concentrates feedback intensified by specialization.

\begin{figure*}[t]
\centering
\includegraphics[width=\textwidth]{
    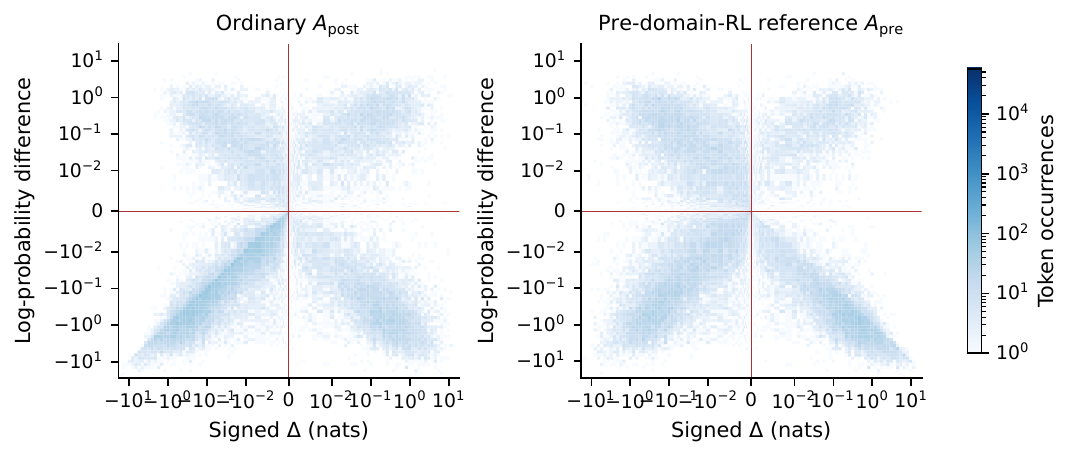
}
\caption{ Token-occurrence densities of signed $\Delta$ against ordinary OPD advantage and pre-domain-RL reference advantage. Axes use symmetric-log scaling, and colors indicate occurrence counts. }
\label{fig:delta-feedback-density}
\end{figure*}

At the opposite tail, $\Delta$ is positive while the mean ordinary advantage remains negative. Specialization increases the teacher's relative preference for these continuations, but the student can still assign them higher probability than the specialist. Their corrective feedback remains supervised. The middle percentiles have mean preference changes and advantages close to zero. Thus, the pre/post-RL comparison determines where supervision is applied, while the teacher--student comparison determines the direction of the retained update. The occurrence-level distributions in Figure~\ref{fig:delta-feedback-density} further illustrate this distinction.

\subsection{Token Frequency and Context Dependence}
\label{app:delta-token-distributions}

For token type or structural category $v$, let $n_v$ count all occurrences and $e_v$ excluded occurrences. We distinguish its share of all positions, $n_v/N$, its share of exclusions, $e_v/E$, and its exclusion rate, $e_v/n_v$. Enrichment is the exclusion rate divided by the overall rate $E/N=4.988\%$. Token identities preserve case and leading whitespace.

\begin{figure}[t]
\centering
\includegraphics[width=0.8\textwidth]{
    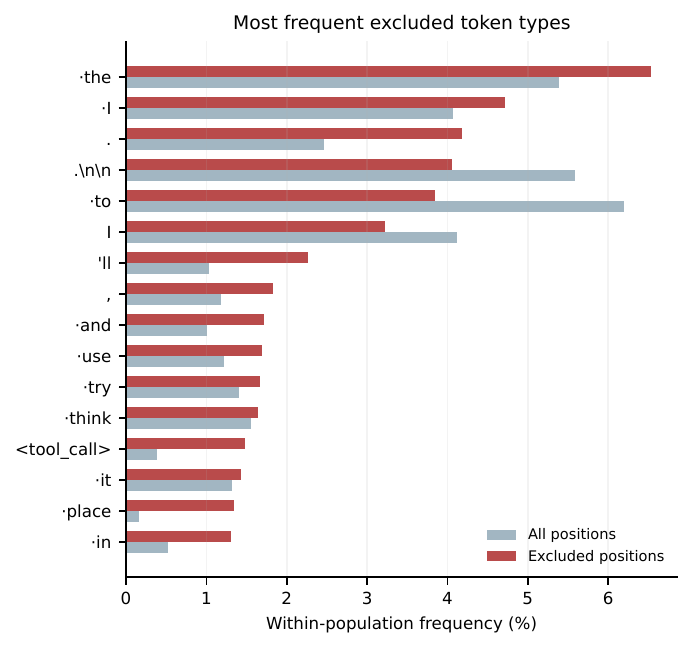
}
\caption{ Frequent excluded token types. Bars show their shares of all valid positions and excluded positions. A centered dot denotes leading whitespace. }
\label{fig:delta-token-frequency}
\end{figure}

Figure~\ref{fig:delta-token-frequency} compares the frequency of token types in all and excluded positions. For example, the leading-space token \texttt{check} is excluded in 99/425 occurrences, whereas \texttt{need} is excluded in 104/2,967. The opening tool-call delimiter is excluded in 128/669 occurrences. These counts distinguish common vocabulary from positions preferentially removed by selection.

\begin{figure}[t]
\centering
\includegraphics[width=0.624\textwidth]{
    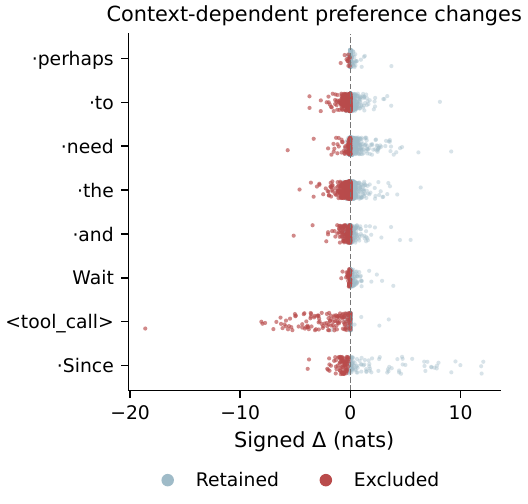
}
\caption{ Signed $\Delta$ for eight token types across their observed histories, ordered by retained fraction. Red occurrences are excluded and blue occurrences retained. Vertical jitter separates occurrences. A centered dot denotes leading whitespace. }
\label{fig:delta-context}
\end{figure}

Preference changes also vary across occurrences of the same token. Of the 547 token types in the excluded tail, 366 also occur in the matched highest-$\Delta$ tail. Figure~\ref{fig:delta-context} shows this variation for eight token types. Selection responds to a token's use in its interaction history, rather than giving each token type a fixed selection status.

\subsection{Illustrative Training Excerpts}
\label{app:delta-colored-cases}

The following excerpts show assistant tokens from Top95 training. Background color represents signed $\Delta$: red indicates decreased preference after domain RL, white indicates little change, and green indicates increased preference. Color saturation is clipped at $|\Delta|=6$. Underlining marks positions excluded by the within-rollout Top95 mask.

\noindent
\opddeltatok{E6A08E}{0}{$-6$}\, \opddeltatok{F2CFC6}{0}{$-3$}\, \opddeltatok{FFFFFF}{0}{$0$}\, \opddeltatok{C7E0D2}{0}{$+3$}\, \opddeltatok{8CC3A4}{0}{$+6$}\par
\smallskip

\begin{nolinenumbers}
\begin{opdcasebox}{1. Tool-call initiation (unsuccessful rollout)}{opdgreen}
\opdrole{Assistant} \opddeltatok{E9AA9A}{1}{<tool\_call>}\par \opddeltatok{FFFFFF}{0}{\{"}\opddeltatok{FFFFFF}{0}{name}\opddeltatok{FFFFFF}{0}{":}\opddeltatok{FFFFFF}{0}{\ "}\opddeltatok{FFFFFF}{0}{alf}\opddeltatok{FFFFFF}{0}{world}\opddeltatok{FFFFFF}{0}{\_step}\opddeltatok{FFFFFF}{0}{",}\opddeltatok{FFFFFF}{0}{\ "}\opddeltatok{FFFFFF}{0}{arguments}\opddeltatok{FFFFFF}{0}{":}\opddeltatok{FFFFFF}{0}{\ \{"}\opddeltatok{FFFFFF}{0}{action}\opddeltatok{FFFFFF}{0}{":}\opddeltatok{FFFFFF}{0}{\ "}\opddeltatok{8CC3A4}{0}{inventory}\opddeltatok{FFFFFF}{0}{"\}\}}\par \opddeltatok{FFFFFF}{0}{</tool\_call>}\opddeltatok{FFFFFF}{0}{<|im\_end|>}

\textit{[Unscored feedback omitted]} \opdrole{Assistant} \opddeltatok{F5DAD3}{1}{<tool\_call>}\par \opddeltatok{FFFFFF}{0}{\{"}\opddeltatok{FFFFFF}{0}{name}\opddeltatok{FFFFFF}{0}{":}\opddeltatok{FFFFFF}{0}{\ "}\opddeltatok{FFFFFF}{0}{alf}\opddeltatok{FFFFFF}{0}{world}\opddeltatok{FFFFFF}{0}{\_step}\opddeltatok{FFFFFF}{0}{",}\opddeltatok{FFFFFF}{0}{\ "}\opddeltatok{FFFFFF}{0}{arguments}\opddeltatok{FFFFFF}{0}{":}\opddeltatok{FFFFFF}{0}{\ \{"}\opddeltatok{FFFFFF}{0}{action}\opddeltatok{FFFFFF}{0}{":}\opddeltatok{FFFFFF}{0}{\ "}\opddeltatok{FBEFEC}{0}{look}\opddeltatok{FFFFFF}{0}{"\}\}}\par \opddeltatok{FFFFFF}{0}{</tool\_call>}\opddeltatok{FFFFFF}{0}{<|im\_end|>}

\textit{[Unscored feedback omitted]} \opdrole{Assistant} \opddeltatok{F4D6CE}{1}{<tool\_call>}\par \opddeltatok{FFFFFF}{0}{\{"}\opddeltatok{FFFFFF}{0}{name}\opddeltatok{FFFFFF}{0}{":}\opddeltatok{FFFFFF}{0}{\ "}\opddeltatok{FFFFFF}{0}{alf}\opddeltatok{FFFFFF}{0}{world}\opddeltatok{FFFFFF}{0}{\_step}\opddeltatok{FFFFFF}{0}{",}\opddeltatok{FFFFFF}{0}{\ "}\opddeltatok{FFFFFF}{0}{arguments}\opddeltatok{FFFFFF}{0}{":}\opddeltatok{FFFFFF}{0}{\ \{"}\opddeltatok{FFFFFF}{0}{action}\opddeltatok{FFFFFF}{0}{":}\opddeltatok{FFFFFF}{0}{\ "}\opddeltatok{FFFFFF}{0}{go}\opddeltatok{FFFFFF}{0}{\ to}\opddeltatok{F9E8E3}{0}{\ cabinet}\opddeltatok{FFFFFF}{0}{\ }\opddeltatok{FFFFFF}{0}{1}\opddeltatok{F1F8F4}{0}{"\}\}}\opddeltatok{F9E9E5}{0}{<|im\_end|>}

\textit{[End of displayed excerpt.]}
\par\smallskip\textbf{Selected scored positions} ($\Delta$, ordinary $A$; mask status):\par
\texttt{<tool\_call>}: $-5.37$, $-12.43$; excluded.\par \texttt{inventory}: $+6.00$, $-7.45$; retained.\par \texttt{cabinet}: $-1.48$, $-4.68$; retained.\par \texttt{<|im\_end|>}: $-1.40$, $-1.37$; retained.\par

\end{opdcasebox}
\end{nolinenumbers}

\noindent
The first excerpt locates selection at entry into a tool call. The opening delimiter is excluded, while much of the enclosed action content remains supervised. The retained token \texttt{inventory} has positive $\Delta$ but negative ordinary advantage, illustrating how a retained position can still receive suppressive OPD feedback.

\medskip

\begin{nolinenumbers}
\begin{opdcasebox}{2. Planning and action preparation (successful rollout)}{opdgreen}
\opdrole{Assistant} \opddeltatok{FFFFFF}{0}{I}\opddeltatok{FFFFFF}{0}{\ need}\opddeltatok{FFFFFF}{0}{\ to}\opddeltatok{FFFFFF}{0}{\ find}\opddeltatok{FFFFFF}{0}{\ a}\opddeltatok{FFFFFF}{0}{\ remote}\opddeltatok{F7FBF9}{0}{control}\opddeltatok{FCF5F3}{1}{\ and}\opddeltatok{E9F4EE}{0}{\ put}\opddeltatok{FFFFFF}{0}{\ it}\opddeltatok{FEFBFA}{0}{\ in}\opddeltatok{FFFFFF}{0}{\ the}\opddeltatok{FFFFFF}{0}{\ sofa}\opddeltatok{FEFAF9}{0}{.}\opddeltatok{FDF9F7}{0}{\ First}\opddeltatok{FFFFFF}{0}{,}\opddeltatok{FFFFFF}{0}{\ I}\opddeltatok{FFFEFE}{0}{\ should}\opddeltatok{F8E3DE}{1}{\ check}\opddeltatok{FFFFFF}{0}{\ where}\opddeltatok{FFFFFF}{0}{\ the}\opddeltatok{FFFFFF}{0}{\ remote}\opddeltatok{FFFFFF}{0}{control}\opddeltatok{FFFFFF}{0}{\ might}\opddeltatok{FFFFFF}{0}{\ be}\opddeltatok{FEFCFB}{0}{.}\opddeltatok{FFFEFE}{0}{\ Since}\opddeltatok{EAF4EE}{0}{\ remote}\opddeltatok{FFFFFF}{0}{controls}\opddeltatok{FFFFFF}{0}{\ are}\opddeltatok{FFFEFE}{0}{\ often}\opddeltatok{CFE6D9}{0}{\ on}\opddeltatok{FFFFFF}{0}{\ tables}\opddeltatok{F2F8F5}{0}{,}\opddeltatok{F9E9E5}{1}{\ in}\opddeltatok{FFFFFF}{0}{\ drawers}\opddeltatok{FFFFFF}{0}{,}\opddeltatok{FFFFFF}{0}{\ or}\opddeltatok{FFFFFF}{0}{\ on}\opddeltatok{FFFFFF}{0}{\ shelves}\opddeltatok{FFFFFF}{0}{,}\opddeltatok{FFFFFF}{0}{\ I}\opddeltatok{FBF0EE}{1}{'ll}\opddeltatok{FFFFFF}{0}{\ start}\opddeltatok{FFFFFF}{0}{\ by}\opddeltatok{FCF4F2}{1}{\ going}\opddeltatok{FFFFFF}{0}{\ to}\opddeltatok{F7E0DA}{1}{\ the}\opddeltatok{FFFDFD}{0}{\ dining}\opddeltatok{FFFEFE}{0}{table}\opddeltatok{FBF2EF}{1}{\ to}\opddeltatok{F6FAF8}{0}{\ check}\opddeltatok{FEFBFA}{0}{\ if}\opddeltatok{FEFAF9}{0}{\ the}\opddeltatok{FFFFFF}{0}{\ remote}\opddeltatok{FFFFFF}{0}{control}\opddeltatok{FFFFFF}{0}{\ is}\opddeltatok{FFFFFF}{0}{\ there}\opddeltatok{FEFFFE}{0}{.}\par
\par
\opddeltatok{FEFFFF}{0}{<tool\_call>}\par \opddeltatok{FFFFFF}{0}{\{"}\opddeltatok{FFFFFF}{0}{name}\opddeltatok{FFFFFF}{0}{":}\opddeltatok{FFFFFF}{0}{\ "}\opddeltatok{FFFFFF}{0}{alf}\opddeltatok{FFFFFF}{0}{world}\opddeltatok{FFFFFF}{0}{\_step}\opddeltatok{FFFFFF}{0}{",}\opddeltatok{FFFFFF}{0}{\ "}\opddeltatok{FFFFFF}{0}{arguments}\opddeltatok{FFFFFF}{0}{":}\opddeltatok{FFFFFF}{0}{\ \{"}\opddeltatok{FFFFFF}{0}{action}\opddeltatok{FFFFFF}{0}{":}\opddeltatok{FFFFFF}{0}{\ "}\opddeltatok{FFFFFF}{0}{go}\opddeltatok{FFFFFF}{0}{\ to}\opddeltatok{FFFFFF}{0}{\ dining}\opddeltatok{FFFFFF}{0}{table}\opddeltatok{FFFFFF}{0}{\ }\opddeltatok{FFFFFF}{0}{1}\opddeltatok{FFFFFF}{0}{"\}\}}\par \opddeltatok{FFFFFF}{0}{</tool\_call>}\opddeltatok{FFFFFF}{0}{<|im\_end|>}

\textit{[Unscored feedback omitted]} \opdrole{Assistant} \opddeltatok{FFFFFF}{0}{I}\opddeltatok{FDFEFE}{0}{\ found}\opddeltatok{FFFFFF}{0}{\ the}\opddeltatok{FFFFFF}{0}{\ remote}\opddeltatok{FFFFFF}{0}{control}\opddeltatok{FFFFFF}{0}{\ on}\opddeltatok{FFFFFF}{0}{\ the}\opddeltatok{FFFFFF}{0}{\ dining}\opddeltatok{FFFFFF}{0}{table}\opddeltatok{F5D9D2}{1}{.}\opddeltatok{FFFFFF}{0}{\ Now}\opddeltatok{F1CBC1}{1}{,}\opddeltatok{FFFFFF}{0}{\ I}\opddeltatok{FFFFFF}{0}{\ need}\opddeltatok{FFFFFF}{0}{\ to}\opddeltatok{FFFFFF}{0}{\ pick}\opddeltatok{FFFFFF}{0}{\ it}\opddeltatok{FFFFFF}{0}{\ up}\opddeltatok{FAFDFB}{0}{\ and}\opddeltatok{FDF7F6}{1}{\ put}\opddeltatok{FFFFFF}{0}{\ it}\opddeltatok{FAFDFB}{0}{\ in}\opddeltatok{FFFFFF}{0}{\ the}\opddeltatok{FFFFFF}{0}{\ sofa}\opddeltatok{FFFFFF}{0}{.}\opddeltatok{FDFEFE}{0}{\ First}\opddeltatok{FFFFFF}{0}{,}\opddeltatok{FFFFFF}{0}{\ I}\opddeltatok{FDFEFD}{0}{'ll}\opddeltatok{FFFFFF}{0}{\ take}\opddeltatok{FFFFFF}{0}{\ the}\opddeltatok{FFFFFF}{0}{\ remote}\opddeltatok{FFFFFF}{0}{control}\opddeltatok{FFFFFF}{0}{\ from}\opddeltatok{FFFFFF}{0}{\ the}\opddeltatok{FFFFFF}{0}{\ dining}\opddeltatok{FFFFFF}{0}{table}\opddeltatok{FFFFFF}{0}{.}\par
\par
\opddeltatok{FFFFFF}{0}{<tool\_call>}\par \opddeltatok{FFFFFF}{0}{\{"}\opddeltatok{FFFFFF}{0}{name}\opddeltatok{FFFFFF}{0}{":}\opddeltatok{FFFFFF}{0}{\ "}\opddeltatok{FFFFFF}{0}{alf}\opddeltatok{FFFFFF}{0}{world}\opddeltatok{FFFFFF}{0}{\_step}\opddeltatok{FFFFFF}{0}{",}\opddeltatok{FFFFFF}{0}{\ "}\opddeltatok{FFFFFF}{0}{arguments}\opddeltatok{FFFFFF}{0}{":}\opddeltatok{FFFFFF}{0}{\ \{"}\opddeltatok{FFFFFF}{0}{action}\opddeltatok{FFFFFF}{0}{":}\opddeltatok{FFFFFF}{0}{\ "}\opddeltatok{FFFFFF}{0}{take}\opddeltatok{FFFFFF}{0}{\ remote}\opddeltatok{FFFFFF}{0}{control}\opddeltatok{FFFFFF}{0}{\ }\opddeltatok{FFFFFF}{0}{1}\opddeltatok{FFFFFF}{0}{\ from}\opddeltatok{FFFFFF}{0}{\ dining}\opddeltatok{FFFFFF}{0}{table}\opddeltatok{FFFFFF}{0}{\ }\opddeltatok{FFFFFF}{0}{1}\opddeltatok{FFFFFF}{0}{"\}\}}\par \opddeltatok{FFFFFF}{0}{</tool\_call>}\opddeltatok{FFFFFF}{0}{<|im\_end|>}

\textit{[Unscored feedback omitted]} \opdrole{Assistant} \opddeltatok{FFFFFF}{0}{I}\opddeltatok{FAFCFB}{0}{\ now}\opddeltatok{FFFFFF}{0}{\ have}\opddeltatok{FFFFFF}{0}{\ the}\opddeltatok{FFFFFF}{0}{\ remote}\opddeltatok{FFFFFF}{0}{control}\opddeltatok{FFFFFF}{0}{\ in}\opddeltatok{FFFFFF}{0}{\ my}\opddeltatok{FFFFFF}{0}{\ inventory}\opddeltatok{FFFFFF}{0}{.}\opddeltatok{FEFFFE}{0}{\ The}\opddeltatok{FFFFFF}{0}{\ next}\opddeltatok{FFFFFF}{0}{\ step}\opddeltatok{FFFFFF}{0}{\ is}\opddeltatok{FFFFFF}{0}{\ to}\opddeltatok{FFFFFF}{0}{\ go}\opddeltatok{FFFFFF}{0}{\ to}\opddeltatok{FFFFFF}{0}{\ the}\opddeltatok{FFFFFF}{0}{\ sofa}\opddeltatok{FFFFFF}{0}{\ and}\opddeltatok{FBEEEB}{1}{\ place}\opddeltatok{FFFFFF}{0}{\ the}\opddeltatok{FFFFFF}{0}{\ remote}\opddeltatok{FFFFFF}{0}{control}\opddeltatok{BEDDCC}{0}{\ inside}\opddeltatok{FFFFFF}{0}{\ it}\opddeltatok{FFFFFF}{0}{.}

\textit{[End of displayed excerpt.]}
\par\smallskip\textbf{Selected scored positions} ($\Delta$, ordinary $A$; mask status):\par
\texttt{put}: $+1.12$, $+0.56$; retained.\par \texttt{check}: $-1.77$, $-1.94$; excluded.\par \texttt{remote}: $+1.12$, $-0.51$; retained.\par \texttt{on}: $+2.48$, $+0.00$; retained.\par \texttt{in}: $-1.38$, $-10.56$; excluded.\par

\end{opdcasebox}
\end{nolinenumbers}

\noindent
The second excerpt connects selection to planning and action preparation. The agent describes a search-and-placement sequence, with different preference changes within the same plan. The retained token \texttt{put} receives positive ordinary feedback, while excluded occurrences such as \texttt{check} and \texttt{in} receive negative feedback. These examples connect the coefficient analysis to concrete uses of language and tool-call boundaries.

%% file: appendix_new/addition_controls.tex
\section{Additional Supervision Experiments}
\label{app:controls}

We examine how supervision coverage, output structure, and feedback direction affect source learning and cross-domain interaction. The ALF-source controls supplement the update-20 comparison in Table~\ref{tab:explorations}. A separate GR-prefix experiment examines whether supervising the beginning of a response changes subsequent mathematical tool use.

\subsection{Learning Dynamics across Retained Fractions}
\label{app:retention-dynamics}

\begin{figure*}[t]
\centering
\includegraphics[width=\textwidth]{
    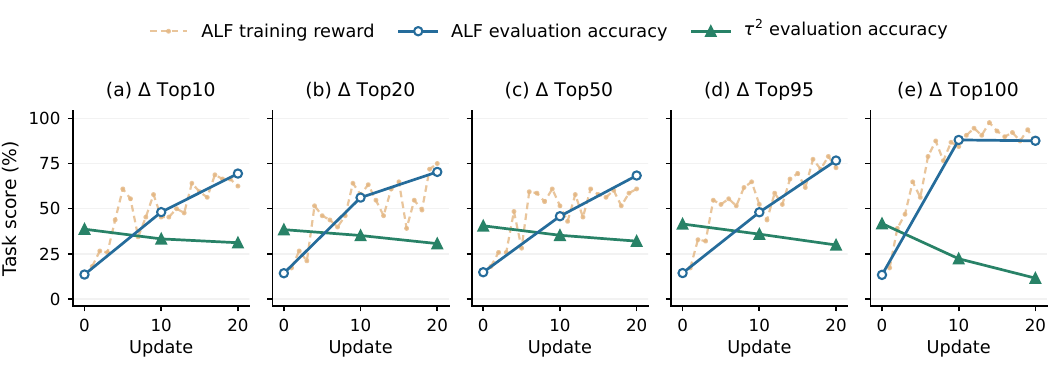
}
\caption{ Learning dynamics across retained fractions. Evaluation curves show ALF and $\tau^2$ accuracy at initialization and updates 10 and 20. Dashed curves show sampled ALF training reward at updates 1--20. All values are percentages. }
\label{fig:retention-learning}
\end{figure*}

Figure~\ref{fig:retention-learning} shows source learning and target performance throughout the selection experiments. Evaluation accuracy is measured at initialization and updates 10 and 20. Training curves report mean ALF task reward on each sampled rollout batch.

The trajectories complement the final-checkpoint comparison by showing how source learning and target degradation develop under different supervision coverage.

\subsection{Structural Masking}
\label{app:structural-controls}

Structural masking tests whether removing supervision at fixed tool-call positions preserves target interaction.

Delimiter masking removes loss on \texttt{<tool\_call>} and \texttt{</tool\_call>}, while retaining the payload and surrounding text. Full-block masking additionally removes the enclosed payload. Its span detector tracks opening-minus-closing counts across the serialized response, so an unmatched opening can extend masking into subsequent assistant positions. Both masks use retained-token normalization (Equation~\ref{eq:opd-masked-loss}).

Table~\ref{tab:explorations} reports their update-20 task outcomes and ask--call co-occurrence. Call construction differs substantially between the two masks: Delimiter produces 4,424 invalid-JSON spans among 15,441 complete $\tau^2$ calls, compared with 33 among 11,905 for Full-block. Fixed structural masking therefore affects call construction as well as the organization of interaction.

\subsection{Negative Masking}
\label{app:positive-only-control}

Negative masking tests whether removing all suppressive feedback reproduces the effect of selective supervision. It uses
\begin{align}
A_t^+
    &=\max(A_{\mathrm{post},t},0),\\
\mathcal L_+
    &=\frac{1}{N}
      \sum_{t\in\mathcal V}
      \ell_t^{\mathrm{OPD}}(\theta;A_t^+),
\end{align}
where $N$ counts all valid assistant positions. The full-token denominator is retained when negative advantages are removed.

Negative masking learns the source task but develops unsupported tool selections on $\tau^2$. At update 20, 7,531 of 8,573 structurally valid, JSON-parseable calls use names absent from the supplied tool schemas. These calls occur in 726 of 800 trajectories and include wrapper names such as \texttt{call\_tool} and \texttt{action}.

\begin{figure*}[t]
\centering
\includegraphics[width=\textwidth]{
    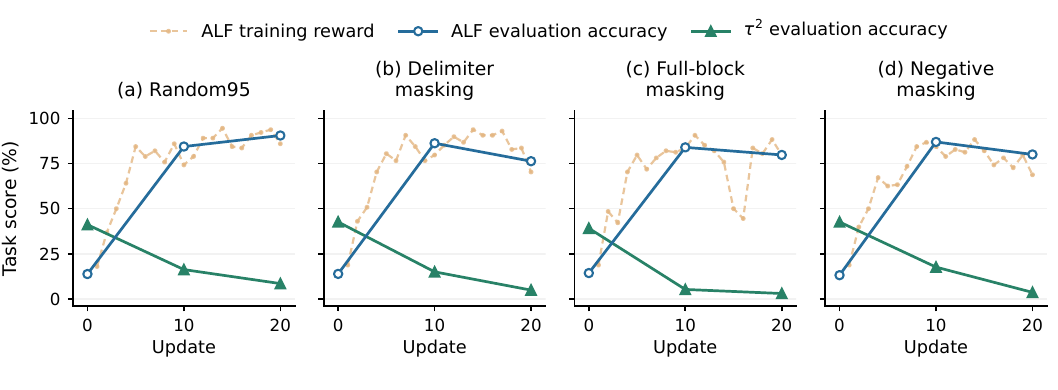
}
\caption{ Learning dynamics of Random95, delimiter masking, full-block masking, and negative masking. Evaluation curves report ALF and $\tau^2$ accuracy; training curves report sampled ALF task reward. }
\label{fig:exploration-learning}
\label{fig:controls}
\label{fig:feedback-explorations}
\end{figure*}

On the fixed Top95 diagnostic contexts, negative masking retains 11.51\% of ordinary absolute coefficient mass. The intervention therefore changes both feedback direction and its retained strength. Figure~\ref{fig:exploration-learning} compares its learning dynamics with the structural and random-selection controls.

\subsection{Ranking Controls and Feedback Strength}
\label{app:ranking-controls-v22}

\paragraph{Selection rules.}
All ranking controls select positions separately within each multi-turn rollout. Among its $N$ valid assistant positions, $K=\lceil\rho N\rceil$ receive ordinary OPD loss, normalized by $K$. The advantage-ranked controls retain the highest raw signed $A_{\mathrm{post},t}$ values, with $\rho=0.95$ or $0.98$ in the main comparison. Absolute-$\Delta$ Top95 retains the highest $|\Delta_t|$ values at $\rho=0.95$, including large changes of either sign. All retain the ordinary signed OPD advantage at selected positions. They do not replace it with the ranking score.

\paragraph{Coefficient diagnostics.}
Table~\ref{tab:w2-coefficients-v22} applies each rule to the same saved Top95 diagnostic contexts and frozen model scores. Excluded mass is the share of absolute ordinary OPD coefficients on excluded positions, computed with full-token normalization. Retained mass additionally accounts for selected-token normalization: within each rollout, its ratio to ordinary OPD is $(1-R_{\mathrm{excl}})N/K$. Both statistics are then averaged over rollouts. They are coefficient diagnostics, not parameter gradient norms; gradient directions and optimizer state are not matched by these calculations.

\begin{table}[!htbp]
\centering\small\setlength{\tabcolsep}{5pt}
\caption{Fixed-context feedback diagnostics. Excluded mass uses full-token normalization; retained mass includes the $N/K$ rescaling. Percentages are rollout-level averages.}
\label{tab:w2-coefficients-v22}
\input{tables/w2_coefficients_v22}
\end{table}

Advantage Top98 removes a comparable amount of coefficient mass on these contexts but selects different positions: its excluded set covers only $19.16\%$ of the signed-$\Delta$ Top95 tail. The resemblance in mass is not maintained by construction during training. On eight saved Advantage Top98 contexts at update 20, the actual mask retains $68.55\%$ of ordinary coefficient mass, whereas applying a virtual signed-$\Delta$ Top95 mask to the same contexts retains $78.36\%$. Absolute-$\Delta$ Top95 removes only $0.021\%$ of coefficient mass on the fixed Top95 contexts. Selected-token normalization then raises retained mass to $105.15\%$ of ordinary OPD. These observations explain why matching the number of retained positions does not match feedback strength, and why the ranking results do not fully separate selection content from optimization strength.

\begin{figure*}[t]
\centering
\includegraphics[width=\textwidth]{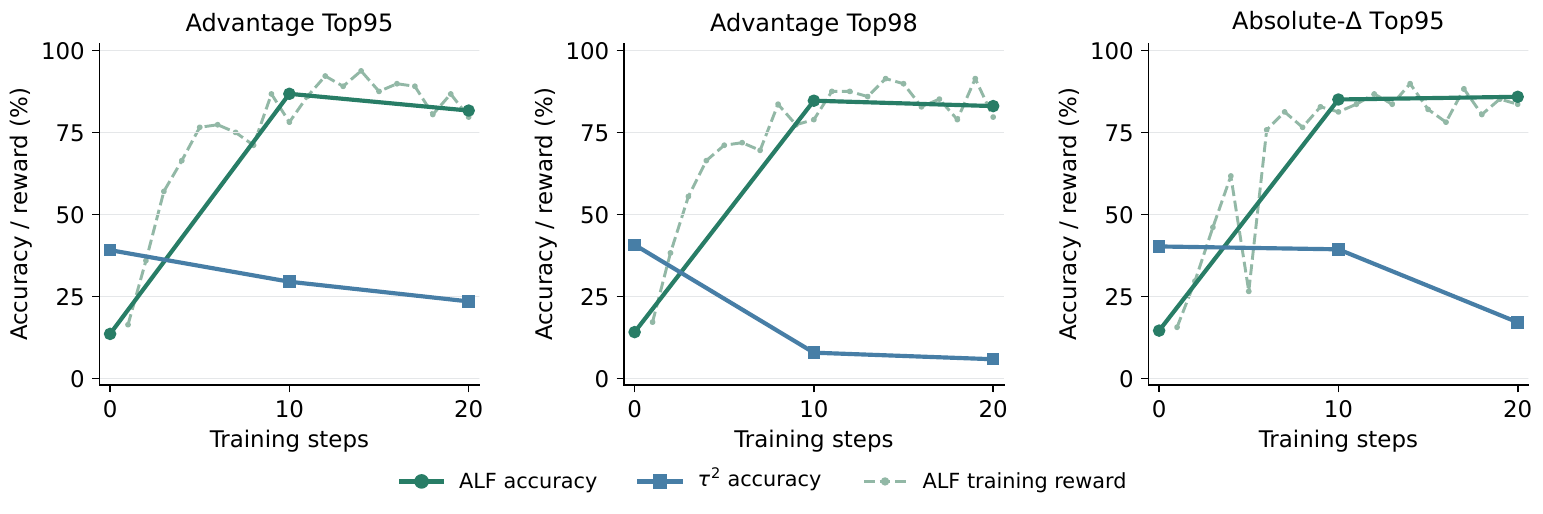} \caption{Ranking-control dynamics over 20 updates. Solid curves show ALF and $\tau^2$ evaluation accuracy at initialization and updates 10 and 20. Light dashed curves show mean ALF training reward on each sampled batch. All values are percentages.}
\label{fig:w2-ranking-dynamics-v22}
\end{figure*}

\paragraph{Learning dynamics.}
Figure~\ref{fig:w2-ranking-dynamics-v22} shows the three added main-text controls over the first 20 updates. Each evaluation contains 1,096 ALF episodes and 800 $\tau^2$ rollouts. The latter cover 100 tasks with eight samples per task. Ask--call co-occurrence uses the number of generated assistant turns containing matched request language as its denominator.

\subsection{GR-to-CI Selective-Supervision Extension}
\label{app:grci-selection-v22}

\paragraph{Teacher, data, and training.}
The extension uses a self-trained Qwen3-30B-A3B GR teacher at its 60-update RL checkpoint and the original pre-RL Qwen3-30B-A3B-Instruct-2507 reference. Both students start from Qwen3-4B-Instruct-2507 and train on 17,916 tool-free GR prompt rows, containing 17,254 unique questions after removal of an exact held-out question overlap. Both conditions use a prompt batch size of 32, four sampled rollouts per prompt, a PPO minibatch size of 32, and an actor learning rate of $10^{-5}$. Prompt and response limits are 8,192 and 32,768 tokens. They share the dual-scorer implementation, with retained fractions of $1.0$ and $0.95$, respectively, and complete 20 student updates.

\paragraph{Paired interfaces.}
CI and GR use identical mathematical questions, answers, and repeat identifiers, with their task-specific system prompts; only CI exposes the code interpreter. Each interface contains 980 expanded samples: 30 AIME2026 questions sampled 16 times and 100 BeyondAIME questions sampled five times. Evaluation uses temperature 1.0 and top-$p$ 1.0. Accuracy is the fraction of correct expanded samples, rather than a uniform average over the 130 unique questions. Tool attempts, executed calls, and correct answer routes are recorded separately.

The update-20 checkpoints are evaluated once more under fixed-checkpoint sampling, with no further training. Table~\ref{tab:grci-selection-v22} reports this companion evaluation.

\begin{figure*}[t]
\centering
\includegraphics[width=\textwidth]{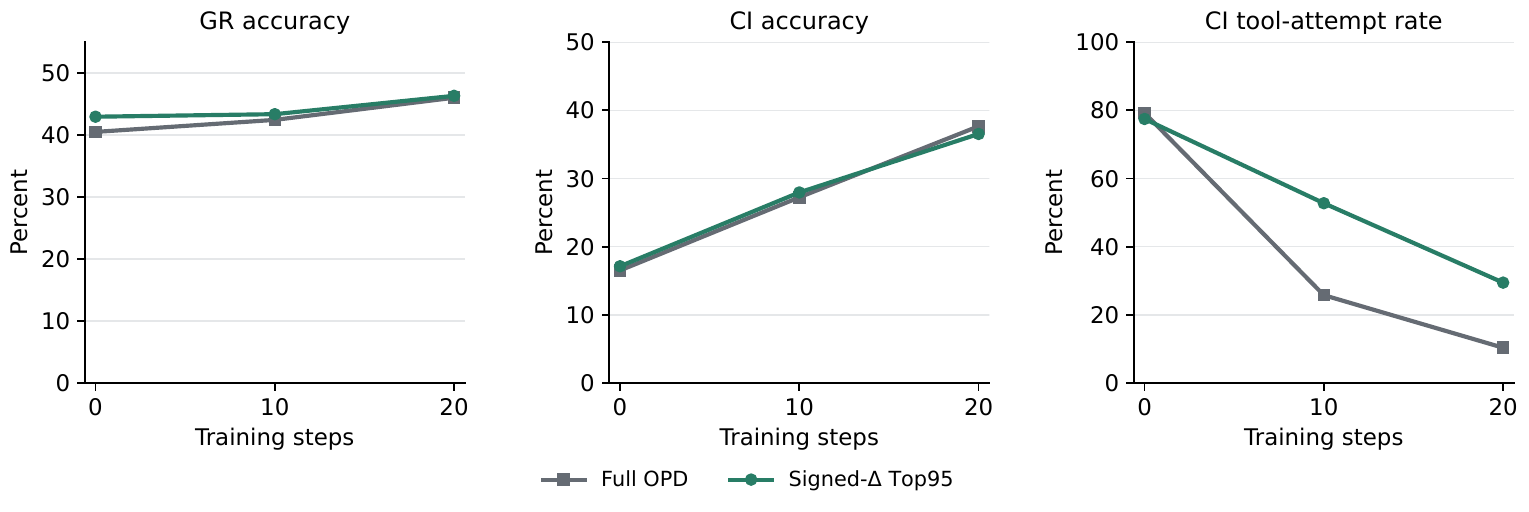} \caption{GR-to-CI learning dynamics. Lines connect the scheduled evaluations at initialization and update 10 with the update-20 companion evaluation reported in Table~\ref{tab:grci-selection-v22}.}
\label{fig:grci-dynamics-v22}
\end{figure*}

The CI response-length-limit rate is $3.78\%$ under Full OPD and $0.10\%$ under Top95; corresponding GR rates are $4.90\%$ and $0.10\%$. These measures describe output-length behavior and are not direct estimates of repetition.

\subsection{GR Prefix Supervision}
\label{app:prefix-control}

The beginning of a mathematical response often introduces the problem representation and solution plan before carrying out detailed derivations or external computation. Inspired by~\citet{ziheng2026less}, we therefore ask whether concentrating OPD supervision on this early prefix can influence subsequent tool use. The motivation is to emphasize strategy formation while reducing supervision on the remainder of the solution trajectory.

The GR-prefix experiment retains valid supervision within the first 1,000 positions of the serialized response tensor:
\begin{equation}
    q_t
    =
    v_t\,\mathbf{1}[t<1000],
\end{equation}
where $v_t$ identifies assistant-generated positions with teacher scores. The position budget is shared across assistant turns. Interleaved observations occupy serialized positions but receive no loss. Selection is based on position, with ordinary OPD feedback applied to the retained tokens.

\begin{figure*}[t]
\centering
\includegraphics[width=\textwidth]{
    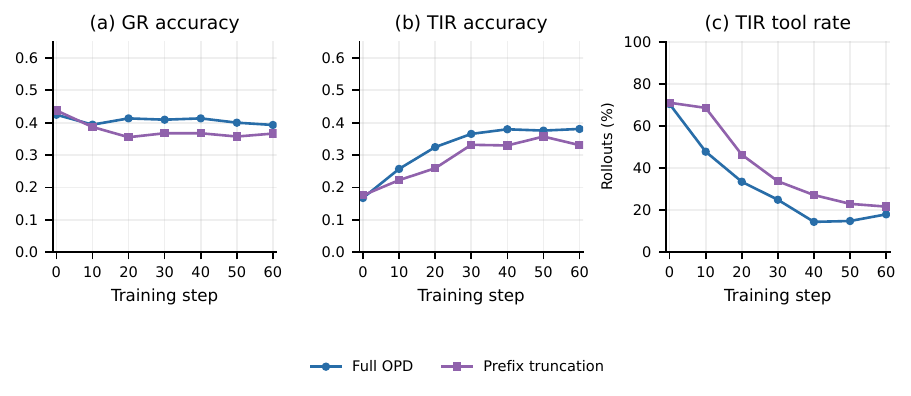
}
\caption{ GR-prefix and full-supervision learning dynamics through 60 updates. Panels report source GR accuracy, tool-enabled mathematical accuracy, and tool-attempt rate. }
\label{fig:prefix}
\end{figure*}

Figure~\ref{fig:prefix} compares prefix supervision with full GR supervision through 60 updates. Under prefix supervision, the tool-enabled mathematical tool-attempt rate falls from 71.0\% to 21.6\%, while accuracy rises from 17.6\% to 33.1\%. Full GR reaches 38.1\% accuracy with an 18.0\% tool-attempt rate at the same update. Source GR accuracy is 36.6\% for the prefix student and 39.3\% for full supervision.

Restricting feedback to the response prefix therefore still produces a substantial shift away from tool use. The higher accuracy under full supervision indicates that feedback on later solution content contributes additional learning.

\subsection{Delta-Based Feedback Exploration}
Inspired by On-Policy Delta Distillation~\citep{heo2026policy}, we also explore directly using pre/post-domain-RL preference differences as distillation feedback. Specifically, we center the delta feedback under the student distribution and retain it when its sign agrees with centered ordinary OPD feedback. Under our multi-turn training setup, this variant exhibits generation degeneration: after 20 updates, both ALFWorld and $\tau^2$ accuracies drop to 0\%, and all 800 $\tau^2$ rollouts reach the response-length limit without generating complete tool calls. This result motivates using $\Delta$ for position selection while retaining the ordinary OPD advantage.

%% file: tables/w2_coefficients_v22.tex
\begin{tabular}{lrr}
\toprule
Rule & Excluded mass (\%) & Retained mass (\%)\\
\midrule
Signed-$\Delta$ Top95 & 41.02 & 62.02\\
Advantage Top95 & 66.88 & 34.83\\
Advantage Top98 & 43.56 & 57.54\\
Absolute-$\Delta$ Top95 & 0.021 & 105.15\\
\bottomrule
\end{tabular}

%% file: appendix_new/training_prompts.tex
\section{Training Prompts and Tool Interfaces}
\label{app:training-prompts}

We reproduce the tool-use instructions and interfaces used in our experiments. The common call format is shown once, followed by the domain-specific prompts. Task instances are omitted; the $\tau^2$ policy excerpts retain their explicit omission markers.

\subsection{Common Tool-Call Format}

CI, ALFWorld, and $\tau^2$ use the following call format.

\begin{nolinenumbers}
\begin{opdcasebox}{Shared tool-call serialization}{opdgreen}

\begin{lstlisting}[style=opdcasecode]
For each function call, return a json object with function name and arguments within <tool_call></tool_call> XML tags:
<tool_call>
{"name": <function-name>, "arguments": <args-json-object>}
</tool_call>
\end{lstlisting}

\end{opdcasebox}
\end{nolinenumbers}

\subsection{Code Interpretation}

The CI prompt allows Python execution for calculation, symbolic manipulation, and verification.

\begin{nolinenumbers}
\begin{opdcasebox}{CI prompt and tool interface}{opdgreen}

\begin{lstlisting}[style=opdcasecode]
You are a math expert. You are given a question and you need to solve it step by step. You should use the `code_interpreter` tool to write and execute Python code for calculations, symbolic math, or result verification when needed. Put your final answer in \boxed{}.
\end{lstlisting}

\begin{lstlisting}[style=opdcasecode]
{
  "type": "function",
  "function": {
    "name": "code_interpreter",
    "description": "Execute Python source code and return its output. Use this tool for exact calculations, symbolic math, enumeration, simulation, and verification.",
    "parameters": {
      "type": "object",
      "properties": {
        "code": {
          "type": "string",
          "description": "The Python code to execute."
        }
      },
      "required": [
        "code"
      ]
    }
  }
}
\end{lstlisting}

\end{opdcasebox}
\end{nolinenumbers}

\subsection{ALFWorld}

The ALFWorld prompt supplies an observation and admissible actions, then asks the agent to reason and execute one listed action.

\begin{nolinenumbers}
\begin{opdcasebox}{ALFWorld prompt and tool interface}{opdgreen}

\begin{lstlisting}[style=opdcasecode]
You are an expert agent operating in the ALFRED Embodied Environment.
At each step you are given your current observation and the list of admissible actions.
Reason step-by-step about the current situation, then take exactly one admissible action by calling the `alfworld_step` tool with that action. The action you pass MUST be copied verbatim from the admissible action list.
\end{lstlisting}

\begin{lstlisting}[style=opdcasecode]
{
  "type": "function",
  "function": {
    "name": "alfworld_step",
    "description": "Take one admissible action in the ALFWorld environment. Choose exactly one action from the current admissible action list.",
    "parameters": {
      "type": "object",
      "properties": {
        "action": {
          "type": "string",
          "description": "The admissible action to take, copied verbatim from the admissible action list."
        }
      },
      "required": [
        "action"
      ]
    }
  }
}
\end{lstlisting}

\end{opdcasebox}
\end{nolinenumbers}

\subsection{Conversational Tool Use}

The three $\tau^2$ domains share turn-level instructions and use their respective domain policies and tool interfaces.

\begin{nolinenumbers}
\begin{opdcasebox}{$\tau^2$: shared interaction instructions}{opdgreen}

\begin{lstlisting}[style=opdcasecode]
You are a customer service agent.
Follow the domain operating rules exactly. On each turn, do exactly one of:
- send a message to the customer; or
- call exactly one available tool.
Do not combine customer-facing text and a tool call in the same turn.
You may call only tools listed in the tool interface supplied to you. Actions described as
customer-side or device-side are performed by the customer, not by you. When such an
action is needed, explain it to the customer and ask the customer to perform it.
Use the done tool only when the task is complete or cannot be completed.
\end{lstlisting}

\end{opdcasebox}
\end{nolinenumbers}

\begin{nolinenumbers}
\begin{opdcasebox}{$\tau^2$ / Airline: policy excerpt and tool fields}{opdgreen}

\begin{lstlisting}[style=opdcasecode]
As an airline agent, you can help users **book**, **modify**, or **cancel** flight reservations. You also handle **refunds and compensation**.

[... other policy clauses omitted ...]

Before taking any actions that update the booking database (booking, modifying flights, editing baggage, changing cabin class, or updating passenger information), you must list the action details and obtain explicit user confirmation (yes) to proceed.

[... other policy clauses omitted ...]

You should only use tools provided in the agent tool interface. Do not attempt to call procedures that are merely described in the policy or in the customer's request. You should only make one tool call at a time, and if you make a tool call, you should not respond to the user simultaneously. If you respond to the user, you should not make a tool call at the same time.
\end{lstlisting}

\textit{Example tool-interface fields (other fields omitted):}

\begin{lstlisting}[style=opdcasecode]
{
  "name": "get_reservation_details",
  "description": "Get the details of a reservation.",
  "required": [
    "reservation_id"
  ]
}
\end{lstlisting}

\end{opdcasebox}
\end{nolinenumbers}

\begin{nolinenumbers}
\begin{opdcasebox}{$\tau^2$ / Retail: policy excerpt and tool fields}{opdgreen}

\begin{lstlisting}[style=opdcasecode]
At the beginning of the conversation, you have to authenticate the user identity by locating their user id via email, or via name + zip code. This has to be done even when the user already provides the user id.

[... other policy clauses omitted ...]

Before taking any action that updates the database (cancel, modify, return, exchange), you must list the action details and obtain explicit user confirmation (yes) to proceed.

[... other policy clauses omitted ...]

You should use only tools provided in the agent tool interface. Do not attempt to call procedures that are merely described in the policy or in the customer's request. You should at most make one tool call at a time, and if you take a tool call, you should not respond to the user at the same time. If you respond to the user, you should not make a tool call at the same time.
\end{lstlisting}

\textit{Example tool-interface fields (other fields omitted):}

\begin{lstlisting}[style=opdcasecode]
{
  "name": "modify_pending_order_payment",
  "description": "Modify the payment method of a pending order. The agent needs to explain the modification detail and ask for explicit user confirmation (yes/no) to proceed.",
  "required": [
    "order_id",
    "payment_method_id"
  ]
}
\end{lstlisting}

\end{opdcasebox}
\end{nolinenumbers}

\begin{nolinenumbers}
\begin{opdcasebox}{$\tau^2$ / Telecom: policy excerpt and tool fields}{opdgreen}

\begin{lstlisting}[style=opdcasecode]
As a telecom agent, you can help users with  **technical support**, **overdue bill payment**, **line suspension**, and **plan options**.

[... other policy clauses omitted ...]

You must first identify the customer.

[... other policy clauses omitted ...]

### Customer-side device procedures
The following procedures are performed by the customer on their own device. They are not
callable tools for the customer service agent. When one of these procedures is needed,
explain the steps to the customer and ask the customer to perform them. Do not emit a
tool call using any of these procedure names.
You must understand these procedures because technical support may require you to guide the customer through them.

[... other policy clauses omitted ...]

#### Troubleshooting Service Problems
##### Airplane Mode
Airplane Mode is a feature that disables all wireless radios, including cellular. If it is enabled, it will prevent any cellular connection.
Ask the customer to check whether Airplane Mode is ON using the status bar or network settings. If it is ON, ask the customer to turn it OFF.
\end{lstlisting}

\textit{Example tool-interface fields (other fields omitted):}

\begin{lstlisting}[style=opdcasecode]
{
  "name": "enable_roaming",
  "description": "Enables international roaming on a line.",
  "required": [
    "customer_id",
    "line_id"
  ]
}
\end{lstlisting}

\end{opdcasebox}
\end{nolinenumbers}

%% file: appendix_new/case_studies_complete.tex
\clearpage
\section{Case Studies}
\label{app:cases}
\label{app:paired-transcripts}

We present two complete comparisons: an urn problem illustrating mathematical solution routes, and a retail task illustrating information retrieval and confirmation before action. Each comparison shows the input, assistant outputs, and returned feedback in their original order. Retail initial user messages are shown separately because their wording differs.

\begingroup \abovedisplayskip=4pt \belowdisplayskip=4pt \abovedisplayshortskip=2pt \belowdisplayshortskip=2pt
\providecommand{\opdcasemath}[1]{\begingroup
\setbox0=\hbox{$\displaystyle #1$}%
\ifdim\wd0>\linewidth
\resizebox{\linewidth}{!}{\box0}%
\else\box0\fi\endgroup}
\renewcommand{\opdrole}[1]{\par\vspace{3pt}\noindent
\textbf{\textcolor{opdgreen}{#1}}\par\nobreak\vspace{1pt}}

\subsection{Enumeration and Algebraic Derivation}

The two models solve the same urn problem. Base uses Python enumeration, while the GR student derives an integer parametrization of the feasible color counts. Both obtain the answer 190.

\begin{nolinenumbers}
\begin{opdcasebox}{Urn problem: Base, correct}{opdgreen}
\opdrole{User (question)} An urn contains $n$ marbles. Each marble is either red or blue, and there are at least $7$ marbles of each color. When $7$ marbles are drawn randomly from the urn without replacement, the probability that exactly $4$ of them are red equals the probability that exactly $5$ of them are red. Find the sum of the five least values of $n$ for which this is possible.

\opdrole{Assistant}

The probability that exactly $k$ of the $7$ marbles drawn are red is given by the hypergeometric distribution. In this distribution, the probability is:

\[
\opdcasemath{\begin{gathered}
P(X = k) = \frac{\binom{r}{k} \binom{n-r}{7-k}}{\binom{n}{7}}
\end{gathered}}
\]

where $r$ is the number of red marbles in the urn, $n-r$ is the number of blue marbles, and $n$ is the total number of marbles.

We are told that the probability of drawing exactly 4 red marbles is equal to the probability of drawing exactly 5 red marbles:

\[
\opdcasemath{\begin{gathered}
\frac{\binom{r}{4} \binom{n-r}{3}}{\binom{n}{7}} = \frac{\binom{r}{5} \binom{n-r}{2}}{\binom{n}{7}}
\end{gathered}}
\]

The denominators are the same, so they cancel out. This simplifies to:

\[
\opdcasemath{\begin{gathered}
\binom{r}{4} \binom{n-r}{3} = \binom{r}{5} \binom{n-r}{2}
\end{gathered}}
\]

We can use properties of binomial coefficients to simplify this equation. Let's expand the terms:
\par - $\binom{r}{4} = \frac{r!}{4!(r-4)!}$
\par - $\binom{n-r}{3} = \frac{(n-r)!}{3!(n-r-3)!}$
\par - $\binom{r}{5} = \frac{r!}{5!(r-5)!}$
\par - $\binom{n-r}{2} = \frac{(n-r)!}{2!(n-r-2)!}$

Substituting these into the equation:

\[
\opdcasemath{\begin{gathered}
\frac{r!}{4!(r-4)!} \cdot \frac{(n-r)!}{3!(n-r-3)!} = \frac{r!}{5!(r-5)!} \cdot \frac{(n-r)!}{2!(n-r-2)!}
\end{gathered}}
\]

We can cancel out the common terms ($r!$ and $(n-r)!$) from both sides:

\[
\opdcasemath{\begin{gathered}
\frac{1}{4!(r-4)!} \cdot \frac{1}{3!(n-r-3)!} = \frac{1}{5!(r-5)!} \cdot \frac{1}{2!(n-r-2)!}
\end{gathered}}
\]

This simplifies to:

\[
\opdcasemath{\begin{gathered}
\frac{1}{(4!(r-4)!)3!(n-r-3)!} = \frac{1}{(5!(r-5)!)2!(n-r-2)!}
\end{gathered}}
\]

Now we can directly compute the values for small $n$ and $r$, ensuring that both $r \geq 7$ and $n - r \geq 7$. Since the problem asks for the sum of the five least values of $n$, we will iterate over possible $n$, and for each $n$, check if there exists an $r$ such that the equation holds and both $r \geq 7$ and $n - r \geq 7$.

Let's write the code to compute these values.
\par\textbf{Tool call: code\_interpreter (decoded code argument)}

\begin{lstlisting}[style=opdcasecode]
from math import comb

def check_probability_equation(n):
    for r in range(7, n - 6):  # r >= 7, n - r >= 7 implies r <= n - 7
        if comb(r, 4) * comb(n - r, 3) == comb(r, 5) * comb(n - r, 2):
            return True
    return False

# Find the five smallest values of n for which the probability condition holds
valid_n_values = []
n = 14  # Starting from n = 14 to satisfy both conditions on r and n - r

while len(valid_n_values) < 5:
    if check_probability_equation(n):
        valid_n_values.append(n)
    n += 1

sum_of_valid_n_values = sum(valid_n_values)
valid_n_values, sum_of_valid_n_values
\end{lstlisting}

\opdrole{Tool response}

\begin{lstlisting}[style=opdcasecode]
([22, 30, 38, 46, 54], 190)
\end{lstlisting}

\opdrole{Assistant}

The five smallest values of $ n $ for which the probability condition holds are $ 22, 30, 38, 46, $ and $ 54 $. The sum of these values is:

\[
\opdcasemath{\begin{gathered}
22 + 30 + 38 + 46 + 54 = 190
\end{gathered}}
\]

Thus, the final answer is $ \boxed{190} $.
\end{opdcasebox}
\end{nolinenumbers}

\begin{nolinenumbers}
\begin{opdcasebox}{Urn problem: GR student, correct}{opdblue}
\opdrole{User (question)} An urn contains $n$ marbles. Each marble is either red or blue, and there are at least $7$ marbles of each color. When $7$ marbles are drawn randomly from the urn without replacement, the probability that exactly $4$ of them are red equals the probability that exactly $5$ of them are red. Find the sum of the five least values of $n$ for which this is possible.

\opdrole{Assistant}

We are told:

\par - An urn contains $ n $ marbles, with at least 7 red and 7 blue marbles.
\par - When 7 marbles are drawn without replacement, the probability of getting exactly 4 red is equal to the probability of getting exactly 5 red.
\par - We are to find the sum of the five least values of $ n $ for which this is possible.

Step 1: Define variables

Let:
\par - $ r $ = number of red marbles in the urn
\par - $ b $ = number of blue marbles in the urn
\par - $ n = r + b $

We are told $ r \geq 7 $, $ b \geq 7 $, and $ n \geq 14 $.

We are drawing 7 marbles without replacement.

The probability of drawing exactly $ k $ red marbles is given by the hypergeometric distribution:

\[
\opdcasemath{\begin{gathered}
P(k) = \frac{\binom{r}{k} \binom{b}{7-k}}{\binom{n}{7}}
\end{gathered}}
\]

We are told:

\[
\opdcasemath{\begin{gathered}
P(4) = P(5)
\end{gathered}}
\]

\[
\opdcasemath{\begin{gathered}
\frac{\binom{r}{4} \binom{b}{3}}{\binom{n}{7}} = \frac{\binom{r}{5} \binom{b}{2}}{\binom{n}{7}}
\end{gathered}}
\]

Since the denominator is the same and nonzero, we can cancel it:

\[
\opdcasemath{\begin{gathered}
\binom{r}{4} \binom{b}{3} = \binom{r}{5} \binom{b}{2}
\end{gathered}}
\]

Step 2: Use identities

Recall:

\[
\opdcasemath{\begin{gathered}
\binom{r}{5} = \binom{r}{r-5},\quad \binom{r}{4} = \binom{r}{r-4}
\end{gathered}}
\]

But we can use the identity:

\[
\opdcasemath{\begin{gathered}
\binom{r}{5} = \binom{r}{4} \cdot \frac{r - 4}{5}
\end{gathered}}
\]

Similarly,

\[
\opdcasemath{\begin{gathered}
\binom{b}{3} = \binom{b}{2} \cdot \frac{b - 2}{3}
\end{gathered}}
\]

Let's plug into the equation:

\[
\opdcasemath{\begin{gathered}
\binom{r}{4} \cdot \binom{b}{3} = \binom{r}{5} \cdot \binom{b}{2}
\end{gathered}}
\]

Substitute:

\[
\opdcasemath{\begin{gathered}
\binom{r}{4} \cdot \left( \binom{b}{2} \cdot \frac{b - 2}{3} \right) = \left( \binom{r}{4} \cdot \frac{r - 4}{5} \right) \cdot \binom{b}{2}
\end{gathered}}
\]

Now divide both sides by $ \binom{r}{4} \cdot \binom{b}{2} $ (assuming they are nonzero, which they are for $ r \geq 5, b \geq 3 $):

\[
\opdcasemath{\begin{gathered}
\frac{b - 2}{3} = \frac{r - 4}{5}
\end{gathered}}
\]

Now solve for $ r $ in terms of $ b $:

\[
\opdcasemath{\begin{gathered}
5(b - 2) = 3(r - 4)
\end{gathered}}
\]

\[
\opdcasemath{\begin{gathered}
5b - 10 = 3r - 12
\end{gathered}}
\]

\[
\opdcasemath{\begin{gathered}
5b + 2 = 3r
\end{gathered}}
\]

\[
\opdcasemath{\begin{gathered}
r = \frac{5b + 2}{3}
\end{gathered}}
\]

We want $ r $ and $ b $ to be integers, $ r \geq 7 $, $ b \geq 7 $, and $ n = r + b $.

So $ r = \frac{5b + 2}{3} $ must be an integer. Therefore, $ 5b + 2 \equiv 0 \pmod{3} $

\[
\opdcasemath{\begin{gathered}
5b \equiv -2 \pmod{3} \Rightarrow 5b \equiv 1 \pmod{3}
\end{gathered}}
\]

But $ 5 \equiv 2 \pmod{3} $, so:

\[
\opdcasemath{\begin{gathered}
2b \equiv 1 \pmod{3} \Rightarrow b \equiv 2 \pmod{3}
\end{gathered}}
\]

So $ b \equiv 2 \pmod{3} $

Let's write $ b = 3k + 2 $, for integer $ k \geq 0 $

Then:

\[
\opdcasemath{\begin{gathered}
r = \frac{5(3k + 2) + 2}{3} = \frac{15k + 10 + 2}{3} = \frac{15k + 12}{3} = 5k + 4
\end{gathered}}
\]

So:
\par - $ b = 3k + 2 $
\par - $ r = 5k + 4 $
\par - $ n = r + b = (5k + 4) + (3k + 2) = 8k + 6 $

We now need:
\par - $ r \geq 7 $ $\rightarrow$ $ 5k + 4 \geq 7 $ $\rightarrow$ $ 5k \geq 3 $ $\rightarrow$ $ k \geq 1 $ (since $ k $ integer)
\par - $ b \geq 7 $ $\rightarrow$ $ 3k + 2 \geq 7 $ $\rightarrow$ $ 3k \geq 5 $ $\rightarrow$ $ k \geq 2 $

So $ k \geq 2 $

Let's compute the values of $ n = 8k + 6 $ for $ k = 2, 3, 4, 5, 6 $, and find the five smallest such $ n $.

\par - $ k = 2 $: $ n = 8(2) + 6 = 16 + 6 = 22 $
\par - $ k = 3 $: $ 24 + 6 = 30 $
\par - $ k = 4 $: $ 32 + 6 = 38 $
\par - $ k = 5 $: $ 40 + 6 = 46 $
\par - $ k = 6 $: $ 48 + 6 = 54 $

So the five least values of $ n $ are:

\[
\opdcasemath{\begin{gathered}
22, 30, 38, 46, 54
\end{gathered}}
\]

Sum them:

\[
\opdcasemath{\begin{gathered}
22 + 30 = 52 \\
52 + 38 = 90 \\
90 + 46 = 136 \\
136 + 54 = 190
\end{gathered}}
\]

[check mark] Final Answer:

\[
\opdcasemath{\begin{gathered}
\boxed{190}
\end{gathered}}
\]
\end{opdcasebox}
\end{nolinenumbers}

The contrast is in the solution procedure: Base searches over candidate totals and red counts, whereas GR reduces the probability equality to a linear relation and enumerates the resulting integer parameter. The final answer agrees across the two routes.

\subsection{Retrieval and Confirmation in Retail}

Both trajectories concern the same annotated payment-change task. Base retrieves user information but does not query the order, then assumes that the requested card is already in use. The CI-trained student retrieves the order state, checks the gift-card balance, obtains confirmation of an alternative card, and modifies the pending order.

\begin{nolinenumbers}
\begin{opdcasebox}{Retail payment change: Base, unsuccessful}{opdrust}
\opdrole{User (initial message)} Hi, I need some quick help. Can you tell me how much balance is left on my gift card? Also, I have a recent order and I'm trying to remember which credit card I used for it---was it Visa, Mastercard, or Amex?
\medskip
\opdrole{Assistant}

To assist you with your gift card balance and the credit card used in your recent order, I'll first need to authenticate your identity. Could you please provide your email address? \opdrole{User}

Sure, my email is isabella.lopez3271@example.com. I'm kind of in a rush, so I'd really appreciate it if we could get to the gift card balance and my order details quickly.

\opdrole{Assistant}

\begin{lstlisting}[style=opdcasecode]
<tool_call>
{
  "name": "find_user_id_by_email",
  "arguments": {
    "email": "isabella.lopez3271@example.com"
  }
}
</tool_call>
\end{lstlisting}

\opdrole{Tool response}

\begin{lstlisting}[style=opdcasecode]
isabella_lopez_6490
\end{lstlisting}

\opdrole{Assistant}

\begin{lstlisting}[style=opdcasecode]
<tool_call>
{
  "name": "get_user_details",
  "arguments": {
    "user_id": "isabella_lopez_6490"
  }
}
</tool_call>
\end{lstlisting}

\opdrole{Tool response}

\begin{lstlisting}[style=opdcasecode]
{"user_id": "isabella_lopez_6490", "name": {"first_name": "Isabella", "last_name": "Lopez"}, "address": {"address1": "710 Sunset Drive", "address2": "Suite 176", "city": "Phoenix", "country": "USA", "state": "AZ", "zip": "85034"}, "email": "isabella.lopez3271@example.com", "payment_methods": {"credit_card_8897086": {"source": "credit_card", "id": "credit_card_8897086", "brand": "visa", "last_four": "8902"}, "gift_card_8245350": {"source": "gift_card", "id": "gift_card_8245350", "balance": 60.0}, "credit_card_8554680": {"source": "credit_card", "id": "credit_card_8554680", "brand": "mastercard", "last_four": "4336"}, "paypal_1621947": {"source": "paypal", "id": "paypal_1621947"}}, "orders": ["#W4923227"]}
\end{lstlisting}

\opdrole{Assistant}

Your gift card balance is \$60.00.

For your recent order (\#W4923227), you used a Visa card ending in 8902.

Let me know if you need any further assistance! \opdrole{User}

Okay, thank you for that. So I want to confirm---if my order already went through on the Visa, can I just leave it that way, because I prefer not to use my other cards? And more importantly, can I apply that \$60 gift card balance to my order now?

\opdrole{Assistant}

Your order is already processed, so no changes to the payment method are needed. Since the order has already been fulfilled, you cannot apply the gift card balance to it directly.

However, if you'd like to use the \$60 gift card balance for a future purchase, you can do so when placing a new order.

Let me know if you need help with anything else! \opdrole{User}

Hmm, that's a bit of a bummer. Since I really don't want to use my Mastercard or Amex because of their balances, is there any way you can still put the gift card toward this order---maybe as a partial payment adjustment or a store credit for that amount? Or would I have to return something and then rebuy it to use the gift card?

\opdrole{Assistant}

Unfortunately, gift card balances cannot be applied as partial payments or store credits to existing orders once they have been fulfilled. The gift card can only be used for future purchases when placing new orders.

To use the balance, you would need to place a new order and apply the gift card during checkout.

If you'd like, I can help you with any other questions or assist with future orders. Let me know how I can support you further!
\end{opdcasebox}
\end{nolinenumbers}

\begin{nolinenumbers}
\begin{opdcasebox}{Retail payment change: CI student, successful}{opdgreen}
\opdrole{User (initial message)} Hi, I need some help quickly---I'm trying to check the balance on my gift card, and I also want to see which credit card I used for my recent order. I might want to use the gift card on that order instead, but if I can't, I'd like to switch the payment to my Visa. Can you look into that for me?
\medskip
\opdrole{Assistant}

I can help you check your gift card balance and see which credit card was used for your recent order. First, I'll need to authenticate your identity to access your profile and order details. Could you please provide your email address or your first name, last name, and zip code? \opdrole{User}

Sure, my email is isabella.lopez3271@example.com. Can you check everything with that?

\opdrole{Assistant}

\begin{lstlisting}[style=opdcasecode]
<tool_call>
{
  "name": "find_user_id_by_email",
  "arguments": {
    "email": "isabella.lopez3271@example.com"
  }
}
</tool_call>
\end{lstlisting}

\opdrole{Tool response}

\begin{lstlisting}[style=opdcasecode]
isabella_lopez_6490
\end{lstlisting}

\opdrole{Assistant}

Thank you, Isabella. I have successfully authenticated your identity. Now, I will retrieve your gift card balance and the payment method used for your recent order. Let me check that information for you.

First, I'll get your user details to see your gift card balance and payment methods. Then, I'll look up your recent order to identify which credit card was used.

Let me proceed with getting your user details.

\begin{lstlisting}[style=opdcasecode]
<tool_call>
{
  "name": "get_user_details",
  "arguments": {
    "user_id": "isabella_lopez_6490"
  }
}
</tool_call>
\end{lstlisting}

\opdrole{Tool response}

\begin{lstlisting}[style=opdcasecode]
{"user_id": "isabella_lopez_6490", "name": {"first_name": "Isabella", "last_name": "Lopez"}, "address": {"address1": "710 Sunset Drive", "address2": "Suite 176", "city": "Phoenix", "country": "USA", "state": "AZ", "zip": "85034"}, "email": "isabella.lopez3271@example.com", "payment_methods": {"credit_card_8897086": {"source": "credit_card", "id": "credit_card_8897086", "brand": "visa", "last_four": "8902"}, "gift_card_8245350": {"source": "gift_card", "id": "gift_card_8245350", "balance": 60.0}, "credit_card_8554680": {"source": "credit_card", "id": "credit_card_8554680", "brand": "mastercard", "last_four": "4336"}, "paypal_1621947": {"source": "paypal", "id": "paypal_1621947"}}, "orders": ["#W4923227"]}
\end{lstlisting}

\opdrole{Assistant}

I now have your user details. Here's what I found:

\par - Gift Card Balance: \$60.00 (ID: gift\_card\_8245350)
\par - Payment Methods:
  - Visa ending in 8902 (ID: credit\_card\_8897086) - Mastercard ending in 4336 (ID: credit\_card\_8554680) - PayPal account (ID: paypal\_1621947)

Next, I'll check the details of your recent order to see which credit card was used. Your recent order ID is \#W4923227. Let me retrieve that order information.

\begin{lstlisting}[style=opdcasecode]
<tool_call>
{
  "name": "get_order_details",
  "arguments": {
    "order_id": "#W4923227"
  }
}
</tool_call>
\end{lstlisting}

\opdrole{Tool response}

\begin{lstlisting}[style=opdcasecode]
{"order_id": "#W4923227", "user_id": "isabella_lopez_6490", "address": {"address1": "710 Sunset Drive", "address2": "Suite 176", "city": "Phoenix", "country": "USA", "state": "AZ", "zip": "85034"}, "items": [{"name": "Bluetooth Speaker", "product_id": "4768869376", "item_id": "7751905257", "price": 321.18, "options": {"color": "red", "battery life": "10 hours", "water resistance": "yes"}}], "status": "pending", "fulfillments": [], "payment_history": [{"transaction_type": "payment", "amount": 321.18, "payment_method_id": "credit_card_8554680"}], "cancel_reason": null, "exchange_items": null, "exchange_new_items": null, "exchange_payment_method_id": null, "exchange_price_difference": null, "return_items": null, "return_payment_method_id": null}
\end{lstlisting}

\opdrole{Assistant}

Here are the details of your recent order:

\par - Order ID: \#W4923227  
\par - Status: Pending  
\par - Item: Bluetooth Speaker (Color: Red, Battery Life: 10 hours, Water Resistance: Yes)  
\par - Price: \$321.18  
\par - Payment Method Used: Mastercard ending in 4336 (ID: credit\_card\_8554680)

You have a gift card balance of \$60.00, which is not sufficient to cover the full order amount of \$321.18. Therefore, you cannot use the gift card to pay for this order.

However, you can switch the payment method to your Visa ending in 8902 (ID: credit\_card\_8897086), which is a valid option for this pending order.

Would you like me to proceed with modifying the payment method of your order to your Visa card? Please confirm with a "yes" if you'd like to proceed.

Note: This change will keep the order in "pending" status, and the original Mastercard payment will be refunded within 5--7 business days. \opdrole{User}

Yes, please switch it to my Visa. Can you confirm once it's done? I'm trying to wrap this up quickly.

\opdrole{Assistant}

I will now modify your order's payment method to your Visa card (ending in 8902). Let me proceed with that action.

\begin{lstlisting}[style=opdcasecode]
<tool_call>
{
  "name": "modify_pending_order_payment",
  "arguments": {
    "order_id": "#W4923227",
    "payment_method_id": "credit_card_8897086"
  }
}
</tool_call>
\end{lstlisting}

\opdrole{Tool response}

\begin{lstlisting}[style=opdcasecode]
{"order_id": "#W4923227", "user_id": "isabella_lopez_6490", "address": {"address1": "710 Sunset Drive", "address2": "Suite 176", "city": "Phoenix", "country": "USA", "state": "AZ", "zip": "85034"}, "items": [{"name": "Bluetooth Speaker", "product_id": "4768869376", "item_id": "7751905257", "price": 321.18, "options": {"color": "red", "battery life": "10 hours", "water resistance": "yes"}}], "status": "pending", "fulfillments": [], "payment_history": [{"transaction_type": "payment", "amount": 321.18, "payment_method_id": "credit_card_8554680"}, {"transaction_type": "payment", "amount": 321.18, "payment_method_id": "credit_card_8897086"}, {"transaction_type": "refund", "amount": 321.18, "payment_method_id": "credit_card_8554680"}], "cancel_reason": null, "exchange_items": null, "exchange_new_items": null, "exchange_payment_method_id": null, "exchange_price_difference": null, "return_items": null, "return_payment_method_id": null}
\end{lstlisting}

\opdrole{Assistant}

The payment method for your order \#W4923227 has been successfully updated to your Visa card ending in 8902.

Here are the updated details:
\par - Payment Method: Visa (ending in 8902)  
\par - Original Mastercard (ending in 4336) has been refunded within 5--7 business days.

All actions are complete. If you need any further assistance, feel free to ask!

You are all set. done
\end{opdcasebox}
\end{nolinenumbers}

The complete interaction locates the difference in the information used before modification. CI checks the pending order and its payment history, then waits for confirmation before changing the payment method. Its mixed message--tool turns coexist with a successful query-and-confirmation sequence.

\endgroup

%% file: references.bib
@article{hinton2015distilling,
  title={Distilling the knowledge in a neural network},
  author={Hinton, Geoffrey and Vinyals, Oriol and Dean, Jeff},
  journal={arXiv preprint arXiv:1503.02531},
  year={2015}
}

@inproceedings{agarwal2024policy,
  title={On-policy distillation of language models: Learning from self-generated mistakes},
  author={Agarwal, Rishabh and Vieillard, Nino and Zhou, Yongchao and Stanczyk, Piotr and Ramos Garea, Sabela and Geist, Matthieu and Bachem, Olivier},
  booktitle={International Conference on Learning Representations},
  volume={2024},
  pages={21246--21263},
  year={2024}
}

@article{song2026survey,
  title={A survey of on-policy distillation for large language models},
  author={Song, Mingyang and Zheng, Mao},
  journal={arXiv preprint arXiv:2604.00626},
  year={2026}
}

@article{li2026every,
  title={Every Coin Has Two Sides: On the Dual Nature of Generalization in On-Policy Distillation of Large Language Models},
  author={Li, Zhaoyi and Kong, Deyang and Wei, Yuan and Yang, Evan and Shen, Ranran and Ihsani, Mahardika Krisna and Yang, Ming and Zhang, Wei and Hao, Chuan and Yang, Jian and others},
  journal={arXiv preprint arXiv:2608.16647},
  year={2026}
}

@article{ma2026mopd,
  title={Mopd: Multi-teacher on-policy distillation for capability integration in llm post-training},
  author={Ma, Wenhan and Wei, Jianyu and Zhao, Liang and Zhang, Hailin and Xiao, Bangjun and Li, Lei and Yang, Qibin and Gao, Bofei and Wang, Yudong and Li, Rang and others},
  journal={arXiv preprint arXiv:2606.30406},
  year={2026}
}

@article{yao2022react,
  title={React: Synergizing reasoning and acting in language models},
  author={Yao, Shunyu and Zhao, Jeffrey and Yu, Dian and Du, Nan and Shafran, Izhak and Narasimhan, Karthik and Cao, Yuan},
  journal={arXiv preprint arXiv:2210.03629},
  year={2022}
}

@article{schick2023toolformer,
  title={Toolformer: Language models can teach themselves to use tools},
  author={Schick, Timo and Dwivedi-Yu, Jane and Dess{\`\i}, Roberto and Raileanu, Roberta and Lomeli, Maria and Hambro, Eric and Zettlemoyer, Luke and Cancedda, Nicola and Scialom, Thomas},
  journal={Advances in neural information processing systems},
  volume={36},
  pages={68539--68551},
  year={2023}
}

@article{barres2025tau,
title = {{$\tau^2$-Bench: Evaluating Conversational Agents in a Dual-Control Environment}},
  author={Barres, Victor and Dong, Honghua and Ray, Soham and Si, Xujie and Narasimhan, Karthik},
  journal={arXiv preprint arXiv:2506.07982},
  year={2025}
}

@article{yang2025qwen3,
  title={Qwen3 technical report},
  author={Yang, An and Li, Anfeng and Yang, Baosong and Zhang, Beichen and Hui, Binyuan and Zheng, Bo and Yu, Bowen and Gao, Chang and Huang, Chengen and Lv, Chenxu and others},
  journal={arXiv preprint arXiv:2505.09388},
  year={2025}
}

@inproceedings{ALFWorld20,
  title ={{ALFWorld: Aligning Text and Embodied
           Environments for Interactive Learning}},
  author={Mohit Shridhar and Xingdi Yuan and
          Marc-Alexandre C\^ot\'e and Yonatan Bisk and
          Adam Trischler and Matthew Hausknecht},
  booktitle = {Proceedings of the International Conference on Learning Representations (ICLR)},
  year = {2021},
  url = {https://arxiv.org/abs/2010.03768}
}

@inproceedings{li2023api,
  title={Api-bank: A comprehensive benchmark for tool-augmented llms},
  author={Li, Minghao and Zhao, Yingxiu and Yu, Bowen and Song, Feifan and Li, Hangyu and Yu, Haiyang and Li, Zhoujun and Huang, Fei and Li, Yongbin},
  booktitle={Proceedings of the 2023 conference on empirical methods in natural language processing},
  pages={3102--3116},
  year={2023}
}

@inproceedings{liu2024uncertainty,
  title={Uncertainty calibration for tool-using language agents},
  author={Liu, Hao and Dou, Zi-Yi and Wang, Yixin and Peng, Nanyun and Yue, Yisong},
  booktitle={Findings of the Association for Computational Linguistics: EMNLP 2024},
  pages={16781--16805},
  year={2024}
}

@inproceedings{arabzadeh2024assessing,
  title={Assessing and verifying task utility in LLM-powered applications},
  author={Arabzadeh, Negar and Huo, Siqing and Mehta, Nikhil and Wu, Qingyun and Wang, Chi and Hassan, Ahmed and Clarke, Charles LA and Kiseleva, Julia},
  booktitle={Proceedings of the 2024 Conference on Empirical Methods in Natural Language Processing},
  pages={21868--21888},
  year={2024}
}

@inproceedings{sullivan2025procedural,
  title={Procedural environment generation for tool-use agents},
  author={Sullivan, Michael and Hartmann, Mareike and Koller, Alexander},
  booktitle={Proceedings of the 2025 Conference on Empirical Methods in Natural Language Processing},
  pages={18555--18573},
  year={2025}
}

@inproceedings{aoki2024first,
  title={First heuristic then rational: Dynamic use of heuristics in language model reasoning},
  author={Aoki, Yoichi and Kudo, Keito and Kuribayashi, Tatsuki and Sone, Shusaku and Taniguchi, Masaya and Sakaguchi, Keisuke and Inui, Kentaro},
  booktitle={Proceedings of the 2024 Conference on Empirical Methods in Natural Language Processing},
  pages={14255--14271},
  year={2024}
}

@article{wei2022chain,
  title={Chain-of-thought prompting elicits reasoning in large language models},
  author={Wei, Jason and Wang, Xuezhi and Schuurmans, Dale and Bosma, Maarten and Xia, Fei and Chi, Ed and Le, Quoc V and Zhou, Denny and others},
  journal={Advances in neural information processing systems},
  volume={35},
  pages={24824--24837},
  year={2022}
}

@article{wang2022self,
  title={Self-consistency improves chain of thought reasoning in language models},
  author={Wang, Xuezhi and Wei, Jason and Schuurmans, Dale and Le, Quoc and Chi, Ed and Narang, Sharan and Chowdhery, Aakanksha and Zhou, Denny},
  journal={arXiv preprint arXiv:2203.11171},
  year={2022}
}

@misc{bytedance_seed_2025_beyondaime,
  author = {{ByteDance-Seed}},
  title = {{BeyondAIME}: Advancing Math Reasoning Evaluation Beyond High School Olympiads},
  year = {2025},
  howpublished = {Hugging Face Dataset},
  url = {https://huggingface.co/datasets/ByteDance-Seed/BeyondAIME}
}

@misc{aime26,
  title={American Invitational Mathematics Examination ({AIME}) 2026},
  author={Zhang, Yifan and {Math-AI Team}},
  year={2026},
  url={https://huggingface.co/datasets/math-ai/aime26}
}

@article{yu2026dapo,
  title={Dapo: An open-source llm reinforcement learning system at scale},
  author={Yu, Qiying and Zhang, Zheng and Zhu, Ruofei and Yuan, Yufeng and Zuo, Xiaochen and Yue, Yu and Dai, Weinan and Fan, Tiantian and Liu, Gaohong and Liu, Lingjun and others},
  journal={Advances in Neural Information Processing Systems},
  volume={38},
  pages={113222--113244},
  year={2026}
}

@article{ziheng2026less,
  title={Less is more: Early stopping rollout for on-policy distillation},
  author={Ziheng, Zhou and Li, Jiaqi and Tang, Huacong and Wu, Ying Nian and Terzopoulos, Demetri},
  journal={arXiv preprint arXiv:2605.27028},
  year={2026}
}

@article{xu2026deepseek,
  title={Deepseek-v4: Towards highly efficient million-token context intelligence},
  author={Xu, Anyi and Lin, Bangcai and Xue, Bing and Wang, Bingxuan and Xu, Bingzheng and Wu, Bochao and Zhang, Bowei and Lin, Chaofan and Dong, Chen and Ling, Chenchen and others},
  journal={arXiv preprint arXiv:2606.19348},
  year={2026}
}

@article{shinn2023reflexion,
  title={Reflexion: Language agents with verbal reinforcement learning},
  author={Shinn, Noah and Cassano, Federico and Gopinath, Ashwin and Narasimhan, Karthik and Yao, Shunyu},
  journal={Advances in neural information processing systems},
  volume={36},
  pages={8634--8652},
  year={2023}
}

@inproceedings{qin2024toolllm,
  title={Toolllm: Facilitating large language models to master 16000+ real-world apis},
  author={Qin, Yujia and Liang, Shihao and Ye, Yining and Zhu, Kunlun and Yan, Lan and Lu, Yaxi and Lin, Yankai and Cong, Xin and Tang, Xiangru and Qian, Bill and others},
  booktitle={International Conference on Learning Representations},
  volume={2024},
  pages={9695--9717},
  year={2024}
}

@inproceedings{zhou2024webarena,
  title={Webarena: A realistic web environment for building autonomous agents},
  author={Zhou, Shuyan and Xu, Frank F and Zhu, Hao and Zhou, Xuhui and Lo, Robert and Sridhar, Abishek and Cheng, Xianyi and Ou, Tianyue and Bisk, Yonatan and Fried, Daniel and others},
  booktitle={International Conference on Learning Representations},
  volume={2024},
  pages={15585--15606},
  year={2024}
}

@inproceedings{liu2024agentbench,
  title={Agentbench: Evaluating llms as agents},
  author={Liu, Xiao and Yu, Hao and Zhang, Hanchen and Xu, Yifan and Lei, Xuanyu and Lai, Hanyu and Gu, Yu and Ding, Hangliang and Men, Kaiwen and Yang, Kejuan and others},
  booktitle={International Conference on Learning Representations},
  volume={2024},
  pages={52989--53046},
  year={2024}
}

@article{debenedetti2024agentdojo,
  title={Agentdojo: A dynamic environment to evaluate prompt injection attacks and defenses for llm agents},
  author={Debenedetti, Edoardo and Zhang, Jie and Balunovic, Mislav and Beurer-Kellner, Luca and Fischer, Marc and Tram{\`e}r, Florian},
  journal={Advances in neural information processing systems},
  volume={37},
  pages={82895--82920},
  year={2024}
}

@article{kojima2022large,
  title={Large language models are zero-shot reasoners},
  author={Kojima, Takeshi and Gu, Shixiang Shane and Reid, Machel and Matsuo, Yutaka and Iwasawa, Yusuke},
  journal={Advances in neural information processing systems},
  volume={35},
  pages={22199--22213},
  year={2022}
}

@article{kirkpatrick2017overcoming,
  title={Overcoming catastrophic forgetting in neural networks},
  author={Kirkpatrick, James and Pascanu, Razvan and Rabinowitz, Neil and Veness, Joel and Desjardins, Guillaume and Rusu, Andrei A and Milan, Kieran and Quan, John and Ramalho, Tiago and Grabska-Barwinska, Agnieszka and others},
  journal={Proceedings of the national academy of sciences},
  volume={114},
  number={13},
  pages={3521--3526},
  year={2017},
  publisher={National Academy of Sciences}
}

@article{lopez2017gradient,
  title={Gradient episodic memory for continual learning},
  author={Lopez-Paz, David and Ranzato, Marc'Aurelio},
  journal={Advances in neural information processing systems},
  volume={30},
  year={2017}
}

@article{yu2020gradient,
  title={Gradient surgery for multi-task learning},
  author={Yu, Tianhe and Kumar, Saurabh and Gupta, Abhishek and Levine, Sergey and Hausman, Karol and Finn, Chelsea},
  journal={Advances in neural information processing systems},
  volume={33},
  pages={5824--5836},
  year={2020}
}

@inproceedings{standley2020tasks,
  title={Which tasks should be learned together in multi-task learning?},
  author={Standley, Trevor and Zamir, Amir and Chen, Dawn and Guibas, Leonidas and Malik, Jitendra and Savarese, Silvio},
  booktitle={International conference on machine learning},
  pages={9120--9132},
  year={2020},
  organization={PMLR}
}

@inproceedings{gu2024minillm,
  title={Minillm: Knowledge distillation of large language models},
  author={Gu, Yuxian and Dong, Li and Wei, Furu and Huang, Minlie},
  booktitle={International Conference on Learning Representations},
  volume={2024},
  pages={32694--32717},
  year={2024}
}

@article{xiao2026mimo,
  title={Mimo-v2-flash technical report},
  author={Xiao, Bangjun and Xia, Bingquan and Yang, Bo and Gao, Bofei and Shen, Bowen and Zhang, Chen and He, Chenhong and Lou, Chiheng and Luo, Fuli and Wang, Gang and others},
  journal={arXiv preprint arXiv:2601.02780},
  year={2026}
}

@article{zeng2026glm,
  title={Glm-5: from vibe coding to agentic engineering},
  author={Zeng, Aohan and Lv, Xin and Hou, Zhenyu and Du, Zhengxiao and Zheng, Qinkai and Chen, Bin and Yin, Da and Ge, Chendi and Huang, Chenghua and Xie, Chengxing and others},
  journal={arXiv preprint arXiv:2602.15763},
  year={2026}
}

@article{team2026kimi,
  title={Kimi k3: Open frontier intelligence},
  author={Team, Kimi and Bai, Tongtong and Bai, Yifan and Bao, Yiping and Cai, Jianfeng and Cai, Xinyuan and Cao, Peizhou and Cao, Yuxuan and Chai, Ziwei and Charles, Y and others},
  journal={arXiv preprint arXiv:2607.24653},
  year={2026}
}

@misc{schulman2020approximating,
  author       = {John Schulman},
  title        = {Approximating {KL} Divergence},
  year         = {2020},
  month        = mar,
  day          = {7},
  howpublished = {\url{http://joschu.net/blog/kl-approx.html}},
  note         = {Blog post}
}

@inproceedings{xue2026simpletir,
  title={Simpletir: End-to-end reinforcement learning for multi-turn tool-integrated reasoning},
  author={Xue, Zhenghai and Zheng, Longtao and Liu, Qian and Li, Yingru and Zheng, Xiaosen and Ma, Zejun and An, Bo},
  booktitle={International Conference on Learning Representations},
  volume={2026},
  pages={8424--8449},
  year={2026}
}

@article{wang2026tcod,
  title={Tcod: Exploring temporal curriculum in on-policy distillation for multi-turn autonomous agents},
  author={Wang, Jiaqi and Zhang, Wenhao and Shi, Weijie and Li, Yaliang and Cheng, James},
  journal={arXiv preprint arXiv:2604.24005},
  year={2026}
}

@article{zhong2026sod,
  title={Sod: Step-wise on-policy distillation for small language model agents},
  author={Zhong, Qiyong and Zheng, Mao and Song, Mingyang and Lin, Xin and Sun, Jie and Jiang, Houcheng and Wang, Xiang and Fang, Junfeng},
  journal={arXiv preprint arXiv:2605.07725},
  year={2026}
}

@article{zhou2026sage,
  title={SAGE-OPD: Selective Agent-Guided Intervention for Multi-Turn On-Policy Distillation},
  author={Zhou, Yuhang and Zhang, Lizhu and Wu, Yifan and Wang, Mingyi and Peng, Bo and Liu, Jiayi and Fan, Xiangjun and Zhao, Zhuokai},
  journal={arXiv preprint arXiv:2606.19659},
  year={2026}
}

@article{wang2026beyond,
  title={Beyond Prompt Count: How Data Shapes Transfer in On-Policy Distillation},
  author={Wang, Jiaxuan and Lyu, Jiafei and Cai, Yuchen and Wu, Siye and Wang, Pengyuan and Liu, Jiashun and Cheng, Xiang and Yang, Kai and Chen, Yangkun and Yang, Saiyong and others},
  journal={arXiv preprint arXiv:2609.37377},
  year={2026}
}

@article{xu2025toward,
  title={Toward large reasoning models: A survey of reinforced reasoning with large language models},
  author={Xu, Fengli and Hao, Qianyue and Shao, Chenyang and Zong, Zefang and Li, Yu and Wang, Jingwei and Zhang, Yunke and Wang, Jingyi and Lan, Xiaochong and Gong, Jiahui and others},
  journal={Patterns},
  volume={6},
  number={10},
  year={2025},
  publisher={Elsevier}
}

@article{shen2026top,
  title={When Top-K Misses the Decision: Tool-Call Drift in Multi-Teacher On-Policy Distillation},
  author={Shen, Jiabin and Chen, Guang and Mao, Chengjun},
  journal={arXiv e-prints},
  pages={arXiv--2607},
  year={2026}
}

@article{heo2026policy,
  title={On-Policy Delta Distillation},
  author={Heo, Byeongho and Hwang, Jaehui and Yun, Sangdoo and Han, Dongyoon},
  journal={arXiv preprint arXiv:2607.15161},
  year={2026}
}

@article{yu2025demystifying,
  title={Demystifying reinforcement learning in agentic reasoning},
  author={Yu, Zhaochen and Yang, Ling and Zou, Jiaru and Yan, Shuicheng and Wang, Mengdi},
  journal={arXiv preprint arXiv:2510.11701},
  year={2025}
}

@misc{wei2025browsecompsimplechallengingbenchmark,
      title={BrowseComp: A Simple Yet Challenging Benchmark for Browsing Agents}, 
      author={Jason Wei and Zhiqing Sun and Spencer Papay and Scott McKinney and Jeffrey Han and Isa Fulford and Hyung Won Chung and Alex Tachard Passos and William Fedus and Amelia Glaese},
      year={2025},
      eprint={2504.12516},
      archivePrefix={arXiv},
      primaryClass={cs.CL},
      url={https://arxiv.org/abs/2504.12516}, 
}

@misc{jimenez2024swebenchlanguagemodelsresolve,
      title={SWE-bench: Can Language Models Resolve Real-World GitHub Issues?}, 
      author={Carlos E. Jimenez and John Yang and Alexander Wettig and Shunyu Yao and Kexin Pei and Ofir Press and Karthik Narasimhan},
      year={2024},
      eprint={2310.06770},
      archivePrefix={arXiv},
      primaryClass={cs.CL},
      url={https://arxiv.org/abs/2310.06770}, 
}

@inproceedings{
hendrycks2021measuring,
title={Measuring Mathematical Problem Solving With the {MATH} Dataset},
author={Dan Hendrycks and Collin Burns and Saurav Kadavath and Akul Arora and Steven Basart and Eric Tang and Dawn Song and Jacob Steinhardt},
booktitle={Thirty-fifth Conference on Neural Information Processing Systems Datasets and Benchmarks Track (Round 2)},
year={2021},
url={https://openreview.net/forum?id=7Bywt2mQsCe}
}

@misc{qian2025toolrlrewardtoollearning,
      title={ToolRL: Reward is All Tool Learning Needs}, 
      author={Cheng Qian and Emre Can Acikgoz and Qi He and Hongru Wang and Xiusi Chen and Dilek Hakkani-Tür and Gokhan Tur and Heng Ji},
      year={2025},
      eprint={2504.13958},
      archivePrefix={arXiv},
      primaryClass={cs.LG},
      url={https://arxiv.org/abs/2504.13958}, 
}

@misc{heyueya2023solvingmathwordproblems,
      title={Solving Math Word Problems by Combining Language Models With Symbolic Solvers}, 
      author={Joy He-Yueya and Gabriel Poesia and Rose E. Wang and Noah D. Goodman},
      year={2023},
      eprint={2304.09102},
      archivePrefix={arXiv},
      primaryClass={cs.CL},
      url={https://arxiv.org/abs/2304.09102}, 
}

@article{jin2026entropy,
  title={Entropy-aware on-policy distillation of language models},
  author={Jin, Woogyeol and Min, Taywon and Yang, Yongjin and Wei, Dennis and Zhou, Yi and Kadhe, Swanand Ravindra and Baracaldo, Nathalie and Lee, Kimin},
  journal={arXiv preprint arXiv:2603.07079},
  year={2026}
}

@article{shen2026geometry,
  title={On the Geometry of On-Policy Distillation},
  author={Shen, Zhennan and Li, Yanshu and Yin, Qingyu and Leong, Chak Tou and Wang, Zhilin and Chen, Yanxu and Han, Rongduo and Lee, Sunbowen and Fung, Yi R},
  journal={arXiv preprint arXiv:2606.07082},
  year={2026}
}

@article{wu2026seed,
  title={SEED: Self-Evolving On-Policy Distillation for Agentic Reinforcement Learning},
  author={Wu, Jinyang and Yang, Shuo and Lu, Zhengxi and Zhang, Fan and Shen, Yuhao and Feng, Lang and Luo, Haoran and Lian, Zheng and Zhang, Shuai and Wen, Zhengqi and others},
  journal={arXiv preprint arXiv:2607.14777},
  year={2026}
}

@article{tan2026atod,
  title={ATOD: Annealed Turn-aware On-policy Distillation for Multi-turn Autonomous Agents},
  author={Tan, Qitai and Zong, Zefang and Li, Yang and Chen, Peng},
  journal={arXiv preprint arXiv:2606.27814},
  year={2026}
}

@article{zheng2026act,
  title={Act First, Reason Later: Accelerating On-Policy Distillation for Multi-Turn Agents via Reference-Conditioned Inverse Dynamics},
  author={Zheng, Zubin and Wu, Jiahao and Zhang, Shaofeng and Zhang, Zhirui and Ong, Yew-Soon and Liu, Shengcai},
  journal={arXiv preprint arXiv:2609.36608},
  year={2026}
}

@article{wu2026lightning,
  title={Lightning opd: Efficient post-training for large reasoning models with offline on-policy distillation},
  author={Wu, Yecheng and Han, Song and Cai, Hai},
  journal={arXiv preprint arXiv:2604.13010},
  year={2026}
}
